\documentclass{article}

    \PassOptionsToPackage{numbers, compress}{natbib}

    \usepackage[preprint]{neurips_2025}

\usepackage[utf8]{inputenc} %
\usepackage[T1]{fontenc}    %
\usepackage{hyperref}       %
\usepackage{url}            %
\usepackage{booktabs}       %
\usepackage{amsfonts}       %
\usepackage{nicefrac}       %
\usepackage{microtype}      %
\usepackage{graphicx}
\usepackage{wrapfig}
\usepackage{subcaption}
\usepackage{float}
\usepackage[tableposition=top]{caption}
\usepackage[table]{xcolor}%
\usepackage{amsmath}
\usepackage{longtable}
\usepackage{enumitem}
\usepackage{algorithm}
\usepackage{algpseudocode}
\usepackage{multicol}
\usepackage{tikz}
\usetikzlibrary{shapes,arrows,positioning,fit,backgrounds,calc,shadows}
\usepackage{diagbox}
\usepackage[breakable]{tcolorbox}

\makeatletter
\g@addto@macro\normalsize{%
  \setlength\abovedisplayskip{2pt}
  \setlength\belowdisplayskip{2pt}
  \setlength\abovedisplayshortskip{2pt}
  \setlength\belowdisplayshortskip{2pt}
}
\makeatother

\floatstyle{plaintop}
\restylefloat{table}

\title{Unlocking Non-Invasive Brain-to-Text}

\author{%
Dulhan Jayalath \quad Gilad Landau \quad \textbf{Oiwi Parker Jones}\\\\
  PNPL\includegraphics[height=1.5\fontcharht\font`\B]{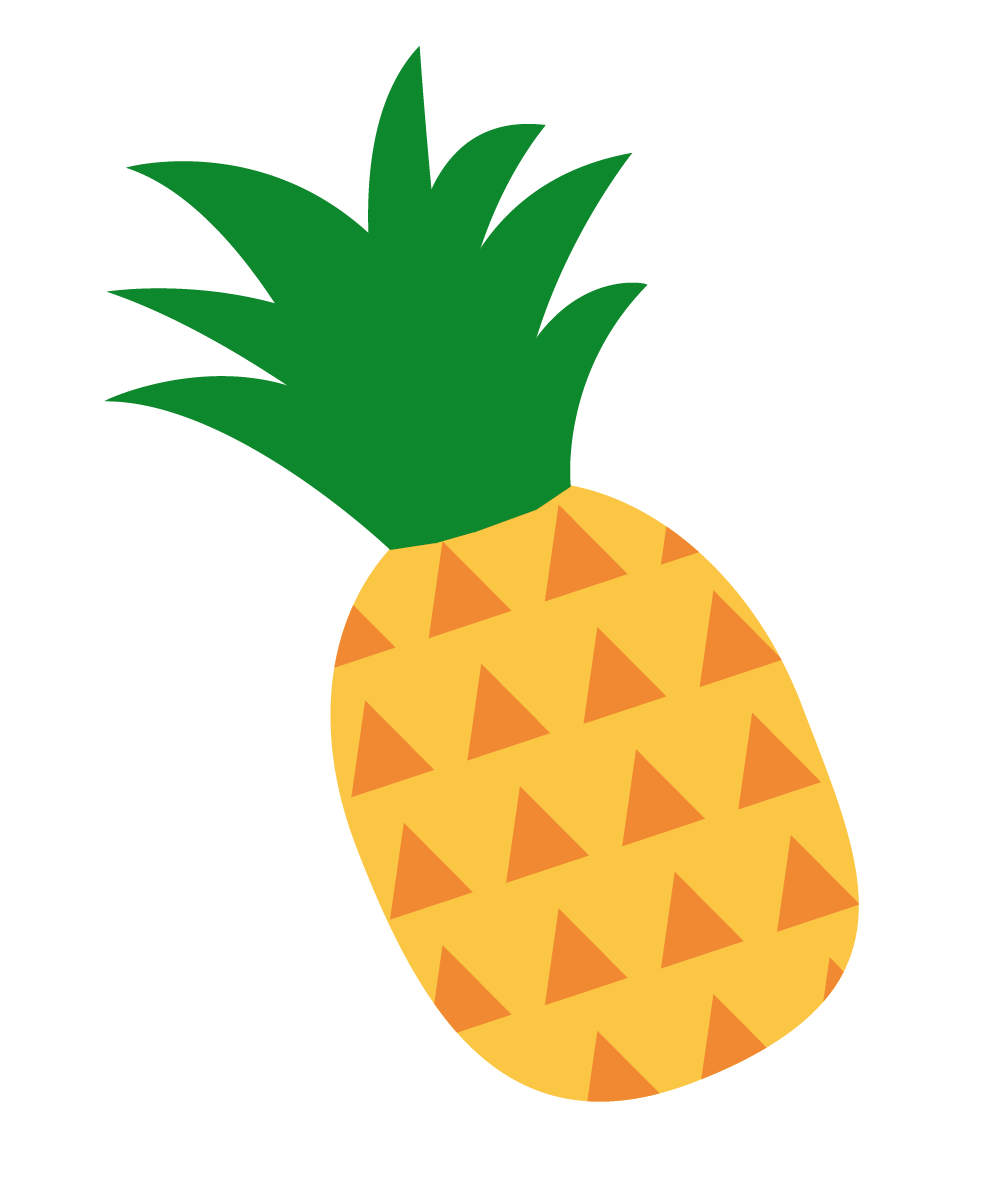}, University of Oxford\\
  \texttt{\{dulhan, oiwi\}@robots.ox.ac.uk} 
}

\begin{document}

\maketitle

\begin{abstract}

  Despite major advances in surgical brain-to-text (B2T), i.e. transcribing speech from invasive brain recordings, non-invasive alternatives have yet to surpass even chance on standard metrics. This remains a barrier to building a non-invasive brain-computer interface (BCI) capable of restoring communication in paralysed individuals without surgery. Here, we present the first non-invasive B2T result that significantly exceeds these critical baselines, \textbf{raising BLEU by $\mathbf{1.4\mathrm{-}2.6\times}$} over prior work. This result is driven by three contributions: (1) we extend recent word-classification models with LLM-based rescoring, transforming single-word predictors into closed-vocabulary B2T systems; (2) we introduce a predictive in-filling approach to handle out-of-vocabulary (OOV) words, substantially expanding the effective vocabulary; and (3) we demonstrate, for the first time, how to scale non-invasive B2T models across datasets, unlocking deep learning at scale and \textbf{improving accuracy by $\mathbf{2.1\mathrm{-}2.3\times}$}. Through these contributions, we offer new insights into the roles of data quality and vocabulary size. Together, our results remove a major obstacle to realising practical non-invasive B2T systems.
\end{abstract}

\begin{wrapfigure}[16]{r}{0.4\textwidth}
    \vspace{-1em}
    \includegraphics[width=0.4\textwidth]{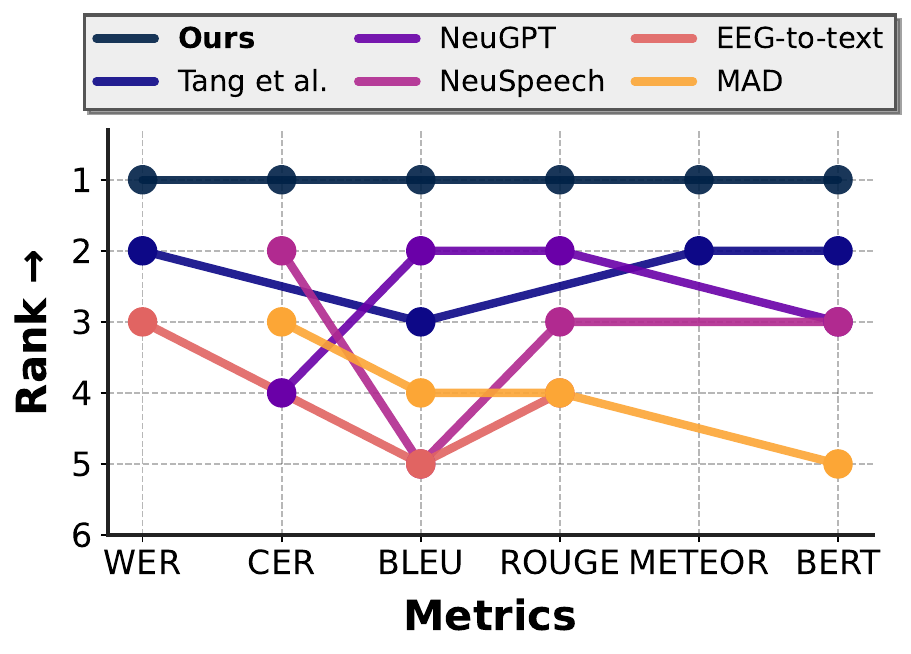}
    \caption{\textbf{Our approach outperforms all non-invasive B2T methods.} Ranks are calculated from absolute improvement over associated random baselines.}
    \label{fig:rank-comp}
\end{wrapfigure}

\section{Introduction}

Transcribing natural language text directly from speech-related neural signals, known as brain-to-text decoding, remains one of neuroscience's most challenging and clinically significant frontiers. Recent breakthroughs showing remarkable accuracy (e.g. \citep{moses2021neuroprosthesis, metzger2023high, willett2023high, card2024accurate, littlejohn2025streaming}) use surgical methods to measure brain activity, implanting electrodes on the surface of, or even inside, the brain. These neurosurgical interventions come with significant inherent risks, including brain infection, haemorrhages, and cognitive side effects, limiting adoption outside of research trials. Instead, safe non-invasive imaging methods, such as EEG and MEG, which measure neural activity with sensors on or around the head, could unlock widespread use. These methods trade surgical risk for lower signal quality as skull and tissue conductivity as well as sensor distance introduce attenuation of the underlying neural signals \citep{ball2009signal}. Despite advances in some classification tasks \citep{jayalath2024brain, defossez2023decoding, dascoli2024decoding}, full brain-to-text reconstruction from non-invasive electrophysiological signals has remained elusive, in spite of numerous attempts \citep{wang2022open, duan2023dewave, yang2024neuspeech, yang2024mad, yang2024neugpt}, with no method convincingly surpassing chance across a set of standard metrics \citep{jo2024eeg}.

\begin{quote}
We present the first non-invasive B2T method to surpass every critical chance baseline across all standard metrics ($p \ll .001$), comprehensively exceeding all prior work (Figure \ref{fig:rank-comp}) and improving over previous BLEU scores by up to $2.6\times$.%
\end{quote}

In reaching this milestone, we address some fundamental challenges. Firstly, we extend individual word prediction \citep{dascoli2024decoding} to contextual sequence decoding, rescoring word predictions according to their surrounding context using a large language model. This enables word predictors to become viable B2T decoders, overcoming the low performance of prior sequence-to-sequence speech decoding strategies \citep{yang2024neuspeech, yang2024mad, yang2024neugpt}. Secondly, word predictors are bound to small closed vocabularies that are hard to scale up, limiting B2T performance. Hence, we propose an in-filling strategy to detect and predict out-of-vocabulary words, enabling open vocabulary transcription from closed-vocabulary decoders.

Although these techniques unlock non-invasive B2T, improving results further requires addressing a data bottleneck. Existing methods can not scale up data because individual speech datasets are small. Moreover, due to drastic differences across datasets, such as hardware, subject anatomy, and cognitive variation, they struggle to combine datasets to achieve meaningful performance improvements in complex tasks (e.g. \citep{jayalath2024brain, gideoni2024non, ridge2024resolving, zhang2023brant, labram2024, wang2024eegpt}). We resolve this fundamental bottleneck by defining standalone dataset performance as a measure of quality. By observing that high quality datasets combine well with other datasets, we selectively pool data to more than double word classification accuracy.

In summary, our work makes three main contributions: (1) we advance non-invasive decoding from single-word prediction to full B2T through contextual LLM rescoring; (2) we transform closed-vocabulary B2T into open-vocabulary decoding with a predictive in-filling strategy; and (3) we enable scaling through our selective dataset pooling framework, improving word classification accuracy by up to $2.3\times$. These advances bring about the first system that surpasses the prerequisite chance baselines for establishing non-invasive B2T and dominates prior work.
Additionally, our ablation studies rigorously validate each component of our approach against controls that invalidated prior methods, establishing a new standard for validating B2T systems.
By demonstrating that non-invasive methods achieve meaningful results, our work challenges the assumption that speech decoding requires surgical imaging, \textit{unlocking} the path for advances in the years to come. %

\section{Related Work}
\label{related-work}

Speech decoding BCIs have progressed along divergent paths: surgical approaches with impressive performance but limited applicability, and non-invasive methods that are safe but must contend with lower signal to noise ratios. This tension between efficacy and accessibility defines the current landscape. In this section, we contextualise our non-invasive work within this rapidly evolving field.

Over the past few years, several breakthroughs have been made in surgical speech BCIs. In 2021, \citet{moses2021neuroprosthesis} developed a B2T neuroprosthesis for a paralysed patient with a 50-word vocabulary. Two years later, \citet{willett2023high} showed similar performance but with a 125,000-word vocabulary. The main innovations of these papers were the recording of high-resolution brain signals and the use of deep neural networks. Following these milestones, subsequent papers have introduced rapid calibration procedures to reduce degradation due to non-stationarities \citep{card2024accurate} and real-time voice synthesis \citep{littlejohn2025streaming}. While these developments are remarkable, surgical approaches carry significant risk of complications and are consequently limited in use outside of controlled clinical trials.

Non-invasive approaches are safe but progress has been limited by noisier signal quality, constraining the most promising work to simpler tasks than B2T. \citet{defossez2023decoding} developed a contrastive learning method that matches pairs of audio and MEG data from speech perception. A preprint by \citet{dascoli2024decoding} extended this method to pairs of word embeddings and MEG data for word classification. While impressive, neither facilitate sequence-level brain-to-text as they either require paired audio \citep{defossez2023decoding} or are bound to small vocabularies and limited by greedy word prediction \citep{dascoli2024decoding}.%

Despite the difficulty, there have been attempts at non-invasive B2T speech decoding. With fMRI, \citet{tang2023semantic} achieved significant results across three participants in speech perception. However, they are not able to decode at word-level granularity as the temporal resolution in fMRI is too low to resolve words, limiting it to a semantic paraphrasing method. With EEG, \citet{wang2022open} and \citet{duan2023dewave} explored open-vocabulary B2T from participants reading text. Subsequent analysis by \citet{jo2024eeg} revealed that performance metrics in these studies were influenced by teacher-forcing during evaluation, and when addressed, performance was comparable to a baseline with random noise inputs. A series of recent MEG-based B2T preprints \citep{yang2024neuspeech, yang2024mad, yang2024neugpt} represent promising directions, though current evaluations lack noise baselines and have not yet surpassed random word selection. Finally, a neighbouring subfield has attempted non-invasive character decoding from brain or muscular activity due to typing or handwriting \citep{sivakumar2024emg2qwerty, levy2025brain}. While important, these are not B2T in the typical sense as they do not decode speech perception, but rather neural activity related to muscle movements. Thus, they require participants to move, making them infeasible for paralysed patients.

The apparent challenge of attaining significant results in non-invasive decoding is largely due to a data bottleneck. Datasets are typically very small, often just several hours, and collected across a few subjects. Thus, reaching deep learning scale necessitates combining datasets, which has yet to be convincingly achieved except for a few cases. These exceptions include brain foundation models \citep{zhang2023brant, labram2024, wang2024eegpt, yang2023biot, kostas2021bendr, ye2023neural, yi2023learning}, particularly \citet{jayalath2024brain} in speech decoding, as well as other approaches like \citet{gideoni2024non} using pooling in source space, and \citet{ridge2024resolving} through adversarial harmonisation. All of these methods show small improvements on simple tasks like classifying speech presence. In more sophisticated tasks, e.g. brain and audio segment matching \citep{defossez2023decoding} or word classification \citep{dascoli2024decoding}, no method has been able to improve performance by combining datasets.

To the best of our knowledge, ours is the first electrophysiological non-invasive B2T approach to: (a) decode sequences beyond chance and other baselines across a range of metrics and (b) show substantial improvements with dataset pooling. In the following section, we describe this B2T method.

\section{Method}

\begin{figure}
    \centering
    \includegraphics[width=1.0\linewidth]{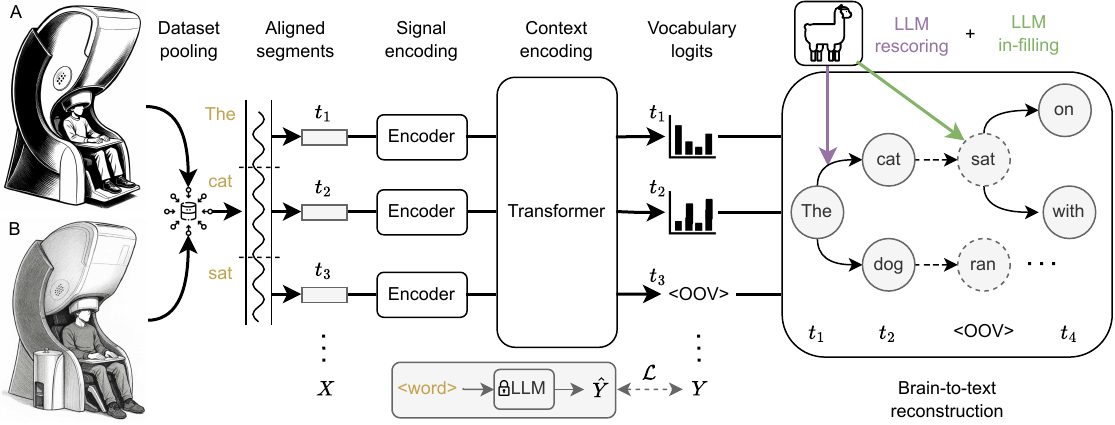}
    \caption{\textbf{Brain-to-text decoding method.} We pool data from multiple heterogeneous datasets (e.g. A and B) and align brain data segments to word onsets. Then, the segments are encoded by a signal encoder, handling dataset differences. A transformer learns the relationships between the encoded latents, embedding them with context. Its outputs are predictions of target word embeddings from a large language model. We map these predictions into logit distributions over the target vocabulary and a beam search with contextual rescoring constructs the highest probability sentence. If we detect out-of-vocabulary words in the stimulus, an in-filling model inserts words into these positions.}
    \label{fig:method}
\end{figure}

Our goal is to take continuous neural data and transcribe continuous text. This requires mapping a recording of brain activity $X \in \mathbb{R}^{s \times T}$, where $s$ is the number of sensors and $T$ is the number of time points in the signal, to a target text $Y \in \mathcal{V}^{N}$ where $\mathcal{V}$ is the vocabulary of the English language and $N$ is the number of words constituting the stimulus text. Figure \ref{fig:method} provides an overview of our method which we describe in detail in the rest of this section.

\subsection{Decoding Words}

We begin by pooling brain recordings from multiple datasets. Then, we segment each brain recording into segments $x_i \in \mathbb{R}^{s\times t}$ of length $t$ time points, where each segment starts at the onset of a word. We call this \textit{aligning} the data. These segments then pass through a signal encoder.

The first stage of the signal encoder resolves differences between segments originating from different datasets and subjects. A spatial attention module \citep{defossez2023decoding} projects data from different datasets with various sensor dimension $s$ into a consistent latent dimension $d_\mathrm{pool}$. Then, we apply a subject-specific linear layer to the latent embedding, maintaining its dimensions. The rest of the signal encoder is a series of dilated convolutions following the structure of standard brain signal encoders \citep{jayalath2024brain, defossez2023decoding, dascoli2024decoding}. These convolutional layers learn a global representation of the neural data, transforming the latent spatial dimension from $d_{\mathrm{pool}}$ to $d_{\mathrm{signal}}$. Finally, to get a time-invariant representation, we mean-pool over the temporal dimension $t$ of the output embedding.

After the signal encoder, we follow \citet{dascoli2024decoding} in explicitly using a transformer to learn contextual relationships between word-aligned neural activity segments. The transformer encoder \citep{vaswani2017attention} learns to process a sequence of $n$ consecutive latent embeddings and outputs $n$ contextual embeddings. We use a contrastive loss (D-SigLIP \citep{dascoli2024decoding}) to learn to match the outputs of the transformer to target word embeddings extracted from the middle layer of a T5 large language model \citep{raffel2020exploring}. We pre-compute the target embeddings from a vocabulary $\mathcal{V}_M$ made up of the top $M$ most frequent words in the text of the dataset, which we call a retrieval set. We map the predicted embedding to a distribution of cosine similarities by measuring its distance to each of the target embeddings in the retrieval set. The model decodes words by predicting the word associated with the target embedding that maximises the cosine similarity to the predicted embedding.

\subsection{Advancing From Words to Sequences}

So far, we have shown only how to predict a limited vocabulary from brain data. In this section, we propose a method for open-vocabulary B2T from closed-vocabulary word prediction.

Our model predicts $n$ cosine similarity distributions, one for each step, simultaneously given a sequence of $n$ neural data segments. We transform each cosine similarity distribution output by our model into a probability distribution $P_{\mathrm{model}}({w}_{i}|\mathbf{x})$ using softmax. If we model each time step independently, the probability of a complete word sequence $\mathbf{w}_{1:n}$ given all neural data segments $\mathbf{x}$ is simply the product of individual word probabilities, leading to the best sequence being given by

\begin{equation}
\label{eq:greedy}
\hat{\mathbf{w}}_{1:n} = \arg\max_{\mathbf{w}_{1:n} \in \mathcal{V}_M^n} \prod_{i=1}^{n} P_{\text{model}}(w_i \mid \mathbf{x}),
\end{equation}

where $\mathcal{V}_M$ is the retrieval set vocabulary. This method is commonly known as \textit{greedy} decoding.

Greedy decoding does not take into account the likelihood of sequences appearing in natural English text. Therefore, we introduce a pre-trained large language model to rescore the sequences based on how representative they are of natural language using the probability the LLM assigns to the sequence. We rescore as part of a beam search where the score of a candidate beam at step $i$ is

\begin{align}
\mathrm{score}(\mathbf{w}_{1:i}|\mathbf{x}, \mathbf{w}_{1:i-1}) &= \mathrm{score}(\mathbf{w}_{1:i-1}|\mathbf{x}, \mathbf{w}_{1:i-2}) \\
&\quad + \log P_{\text{model}}({w}_{i} \mid \mathbf{x}) \\
&\quad + \lambda\log P_{\text{rescorer}}(\mathbf{w}_{1:i})
\end{align}

and the final predicted sequence is chosen as

\begin{equation}
\label{eq:beam}
\hat{\mathbf{w}}_{1:n} = \arg\max_{\mathbf{w}_{1:n} \in \mathcal{B}} \left[ \mathrm{score}(\mathbf{w}_{1:n}|\mathbf{x}, \mathbf{w}_{1:n-1}) \right],
\end{equation}

where $\mathcal{B}$ is the set of beam candidates, $\lambda$ is a weight balancing the encoding model and LLM contributions, and $P_{\text{rescorer}}(\mathbf{w}_{1:n})$ denotes the probability of the word sequence under the LLM. In our experiments, $P_{\text{rescorer}}$ comes from Llama 3.2-1B \citep{llama322024}. Related methods \citep{willett2023high} have used phoneme-level language models to rescore predictions with a Viterbi search \citep{viterbi1967}, an exact and optimal decoding algorithm. The computational cost of LLMs and word-level decoding over long sequences makes this computationally infeasible, so we use beam search as an approximate alternative.

Word-level decoding is limited by closed vocabularies. While increasing the vocabulary permits decoding more words, larger vocabularies reduce decoding performance (Appendix \ref{app:vocab-coverage}). We propose taking advantage of performant vocabularies while compensating for out-of-vocabulary (OOV) words by detecting them and in-filling the missing word using context.

To detect OOV words, we hypothesise that the model will be less confident when presented with neural responses to words it has not seen. As the encoder's output probability distribution over words $\mathbf{p}_i \in \mathbb{R}^{|\mathcal{V}_M|}$ describes the model's confidence, we extract feature vectors from the trained encoder $\mathbf{f}_i = [\mathbf{p}_i, \phi(\mathbf{p}_i)]$, where $\phi(\mathbf{p}_i)$ is a set of distribution statistics, e.g. entropy. We fit a classifier to these features and estimate the probability a position is OOV, i.e. $P(w_i \notin \mathcal{V}_M|\mathbf{f}_i)$. This classifier selects positions to in-fill during inference. We provide further details on OOV detection in Appendix \ref{app:oov-pred}. Next, we describe several possible in-filling strategies.

During beam decoding, we incrementally construct candidate word sequences $\mathbf{w}_{1:i}$. At each decoding step $i$, the model proposes the top $K$ next-word candidates from its retrieval set vocabulary $\mathcal{V}_M$. For each beam hypothesis, if a proposed word $w_i \notin \mathcal{V}_M$, it is replaced in-place by an LLM completion conditioned on the current hypothesis.

Specifically, each beam is updated with the next word $w_i$ chosen as

\begin{equation}
\label{eq:beam-fill}
w_i =
\begin{cases}
\arg\max_{w_i \in \mathcal{V}_M} \mathrm{score}(\mathbf{w}_{1:i}|\mathbf{x}, \mathbf{w}_{1:i-1}), & \text{if } w_i \in \mathcal{V}_M \\
\arg\max_{w_i \in \mathcal{V}_{\text{filler}}} P_{\text{filler}}(w_i \mid \mathbf{w}_{1:i-1}), & \text{if } w_i \notin \mathcal{V}_M
\end{cases}
\end{equation}

where $P_{\mathrm{filler}}$ is the LLM completion probability. Here, the full beam search proceeds with these substituted hypotheses and we select the best hypothesis from the beam set as before (Equation \ref{eq:beam}). Note that $P_{\text{filler}}(w_i \mid \mathbf{w}_{1:i-1})$ is not trivial to calculate from a typical LLM as they operate at the token level rather than the word level. To get word probabilities from token-level probabilities, we use a further token-level beam search over the LLM predictions, where the beam hypothesis is a word hypothesis. While the LLM used for in-filling could be different to the one used for rescoring, the weights are shared in our experiments.

In-filling during the beam search has two limitations. First, it requires another beam search to resolve token-level probability distributions into word hypotheses. Second, it uses only knowledge of prior words in the sequence and not future words. In response, we design an in-context LLM method that in-fills missing positions only after generating the entire best beam sequence. In this method, we perform the full beam search as in Equation \ref{eq:beam} while inserting \texttt{<UNK>} in the positions of out-of-vocabulary words. We then prompt an LLM with the best beam sequence

\begin{equation}
\label{eq:inctxt-infill}
W = \{w_0, w_1, \ldots, w_{n}\} \text{ where } w_i \in \mathcal{V}_M \cup \{\texttt{<UNK>}\},
\end{equation}

and instructions to fill in the missing positions and output the complete sequence. As the context now features all the known words in the sequence, in-filling makes use of both prior and future context. While this could equally be possible with a bidirectional LLM (e.g. BERT \citep{devlin2019bert}), doing this in-context allows direct in-filling of words rather than indirectly via tokens. 

This method, too, has limitations. While in-filling during beam search does not use information from later in the candidate sequence, in-context in-filling does not rescore while taking into account words that could fill missing positions. These limitations can be addressed by both rescoring and in-filling at the same time. To do this, we prompt an LLM as before, but instead of beam rescoring first, we let the LLM choose from the five highest probability words. Thus, rather than a beam sequence, our prompt contains a sequence of quintuples of words and their probabilities

\begin{align}
\label{eq:inctxt-transcribe}
P = \{p_0, p_1, \ldots, p_{n}\} \text{ where } p_i \in \mathcal{C}_5 \cup \{\texttt{<UNK>}\} \\
\quad \text{and} \hspace{1em} \mathcal{C}_5 = \{(w_{i1}, \rho_{i1}), \ldots, (w_{i5}, \rho_{i5}) \mid w_{ij} \in V, \rho_{ij} \in [0,1]\}.
\end{align}

Here, $p_i$ is a quintuple consisting of pairs of words $w_{ij}$ and their probabilities $\rho_{ij}$. The LLM selects the best word for known positions based on its notion of sentence coherency, and fills in the missing word for unknown positions, those marked with \texttt{<UNK>}, outputting a complete sequence.

In both LLM-based methods, we instruct the LLM to output an enumerated dictionary in the format \texttt{\{1: $w_1$, 2: $w_2$, ..., $n$: $w_n$\}}. We find that this constrained format leads to more robust responses with the correct number of outputs $n$, which we attribute to the enumerated keys. Additionally, LLM-based methods can use modes that allow self-reflective chain-of-thought to more deeply reason about rescoring and in-filling. We leverage this in our experiments using Claude Sonnet 3.7 thinking \citep{claude2025}. Our prompts are in Appendix \ref{app:prompts}.

\section{Experiments}

\textbf{Datasets}\quad Our main dataset is LibriBrain \citep{landau2025competition, ozdogan2025libribrain}, a 50-hour MEG dataset collected from a subject listening to Sherlock Holmes audiobooks. We also use three auxiliary datasets for pooling and other experiments: \citet{armeni202210}, containing MEG recordings of 3 subjects listening to 10 hours each of stories, \citet{gwilliams2023introducing} with MEG from 27 subjects, each listening to 4 short stories, and \citet{broderick2018electrophysiological}, providing EEG from 19 subjects, each listening to 20 short speech segments.

\textbf{Methods}\quad \textit{Greedy} describes taking the highest probability words (Equation \ref{eq:greedy}), \textit{beam} refers to LLM rescoring (Equation \ref{eq:beam}), \textit{+fill} indicates in-filling words while rescoring (Equation \ref{eq:beam-fill}), \textit{+IC fill} denotes in-filling words in-context after rescoring (Equation \ref{eq:inctxt-infill}), and \textit{IC transcribe} is both in-context rescoring and in-filling (Equation \ref{eq:inctxt-transcribe}). We provide two kinds of random baselines which we describe in Appendix \ref{app:random-baselines}. When specifying \textit{OOV-D}, we use our OOV detector to select in-filling positions.

\textbf{Evaluation}\quad We quote six brain-to-text metrics covering a broad range of sentence decoding facets. Word error rate (WER) and character error rate (CER) are coarse- and fine-grained measures of exact match decoding accuracy. As prior B2T work measure n-gram overlap, we also quote BLEU-1 \citep{papineni2002bleu}, ROUGE-1F \citep{lin2004rouge}, and METEOR \citep{banerjee2005meteor} as these are standard \citep{tang2023semantic, jo2024eeg, yang2024neugpt}. Finally, we quote BERTScore \citep{bertscore2020} as a measure of semantic similarity. When we do not decode sequences, we provide top-10 balanced word classification accuracy. This is a macro average equivalent to calculating the accuracy for each word in the vocabulary and taking the mean of these accuracies. In tables, we bold the best result and underline the second-best result in a column. Empty cells indicate metrics not reported in the original paper. We always quote the mean and standard error.

\subsection{Realising Non-Invasive Brain-To-Text}

\begin{table}
    \centering
    \begin{tabular}{l|cccccc}
        \toprule
        \textbf{Method} & WER $\downarrow$ & CER $\downarrow$ & BLEU $\uparrow$ & ROUGE $\uparrow$ & METEOR $\uparrow$ & BERT $\uparrow$ \\
        \midrule
        \citet{tang2023semantic} & $0.93$ & & $.24$ & & $\mathbf{.17}$ & $\mathbf{.81}$ \\
        EEG-to-text \citep{wang2022open, duan2023dewave} & $1.00$ & & $.14$ & $.12$ & & \\
        \textbf{Ours} (best) & \cellcolor{green!25}$\mathbf{0.88}$ & \cellcolor{green!25}$\mathbf{0.68}$ & \cellcolor{green!25}$\mathbf{.25}$ & \cellcolor{green!25}$\mathbf{.26}$ & \cellcolor{green!25}$.15$ & \cellcolor{green!25}$\mathbf{.81}$ \\
        \quad Greedy & $\mathbf{0.88}$ & $0.80$ & $.21$ & $.21$ & $.12$ & $.72$ \\
        \quad Beam+fill & $0.91$ & $\mathbf{0.68}$ & $\mathbf{.25}$ & $\mathbf{.26}$ & $\underline{.15}$ & $.80$ \\
        \quad Beam+IC fill & $\underline{0.90}$ & $0.71$ & $.24$ & $.24$ & $\underline{.15}$ & $\mathbf{.81}$ \\
        \midrule
        \midrule
        \textit{Delta to random} \\
        \citet{tang2023semantic} & $\underline{-.03}$ &  & $+.05$ & & $\underline{+.04}$ & $\underline{+.02}$ \\
        EEG-to-text \citep{wang2022open, duan2023dewave} & $+.00$ &  & $+.00$ & $+.00$ &  &  \\
        \textbf{Ours} (best) & \cellcolor{green!25}$\mathbf{-.12}$ & \cellcolor{green!25}$\mathbf{-.19}$ & \cellcolor{green!25}$\mathbf{+.18}$ & \cellcolor{green!25}$\mathbf{+.18}$ & \cellcolor{green!25}$\mathbf{+.09}$ & \cellcolor{green!25}$\mathbf{+.04}$ \\
        
        \bottomrule
    \end{tabular}
    \caption{\textbf{Our method improves non-invasive B2T over established alternatives.} We quote the mean over the three subjects of \citet{tang2023semantic} and the best results for \citet{wang2022open} from a replication study without teacher-forcing \citep{jo2024eeg}. \textit{Delta to random} shows the difference between a method and its associated random baseline.}
    \label{tab:main-results}
\end{table}

On raw brain-to-text performance, we outperform all of the previously best reported non-invasive methods (Table \ref{tab:main-results}). These results establish a new state-of-the-art in non-invasive speech decoding across a range of metrics. We provide annotated decoding examples and more in Appendix \ref{app:examples}.

As prior work uses different experimental protocols, the table also provides the difference between the best method and any quoted null or random baseline (\textit{delta to random}). Here, our approach shows a substantially stronger improvement than any previous work. When compared to prior work in EEG \citep{wang2022open, duan2023dewave}, we achieve significantly better results across all metrics. This is unsurprising as the EEG-to-text methods are no better than chance \citep{jo2024eeg}. Compared to the state-of-the-art fMRI method \citep{tang2023semantic}, ours equals or outperforms it across nearly all metrics.

\begin{table}[t]
    \centering
    \begin{tabular}{l|llllll}
    \toprule
    \textbf{Method} & WER $\downarrow$ & CER $\downarrow$ & BLEU $\uparrow$ & ROUGE $\uparrow$ & METEOR $\uparrow$ & BERT $\uparrow$ \\
    \midrule
    Random \citep{yang2024neugpt} & & $.87$ & $.06$ & $.07$ & & $.84$ \\
    Random (ours) & $1.00_{\pm .0003}$ & $.91_{\pm .004}$ & $.04_{\pm .001}$ & $.04_{\pm .002}$ & $.04_{\pm .001}$ & $.77_{\pm .0003}$ \\
    \midrule
    NeuSpeech \citep{yang2024neuspeech} & & $.77$ & $.05$ & $.08$ & & $\mathbf{.84}$ \\
    MAD \citep{yang2024mad} & & $.90$ & $.07$ & $.07$ & & $.83$ \\
    NeuGPT \citep{yang2024neugpt} & & $1.0$ & $.13$ & $.13$ & & $\mathbf{.84}$ \\
    \midrule
    \textit{Ours} \\
    \quad Beam+fill & $\underline{0.97}_{\pm .001}$ & $\mathbf{.71}_{\pm .001}$ & $\mathbf{.18}_{\pm .003}$ & $\mathbf{.19}_{\pm .003}$ & $\mathbf{.12}_{\pm .001}$ & $.79_{\pm .001}$ \\
    \quad Beam+IC fill & $0.98_{\pm .001}$ & $.78_{\pm .003}$ & $\mathbf{.18}_{\pm .003}$ & $.18_{\pm .003}$ & $\mathbf{.12}_{\pm .002}$ & $.80_{\pm .001}$ \\
    \textit{Ours (pooled)} \\
    \quad Beam+fill & $\mathbf{0.95}_{\pm .001}$ & $\underline{.72}_{\pm .004}$ & $\mathbf{.18}_{\pm .003}$ & $\mathbf{.19}_{\pm .003}$ & $.11_{\pm .002}$ & $.78_{\pm .003}$ \\
    \quad Beam+IC fill & $\mathbf{0.95}_{\pm .003}$ & $.82_{\pm 0.009}$ & $.17_{\pm 0.004}$ & $.17_{\pm 0.005}$ & $.11_{\pm 0.002}$ & $.80_{\pm 0.001}$ \\
    \bottomrule
    \end{tabular}
    \caption{\textbf{Our method surpasses all prior MEG brain-to-text approaches.} As no code is available to reproduce NeuGPT \citep{yang2024neugpt}, we validate and test with the same data and splits and quote their results in this table, including their baselines \citep{yang2024neuspeech, yang2024mad}. We also provide two random selection baselines, one quoted from \citet{yang2024neugpt} and our own, which shows random performs worse in our experimental setup. \textit{Pooled} indicates the model was additionally trained jointly with LibriBrain and Armeni.}
    \label{tab:gwi-yang}
\end{table}

While there are no established works in B2T from MEG, there are a series of recent preprints \citep{yang2024neuspeech, yang2024mad, yang2024neugpt} which our methods also convincingly outperform (Table \ref{tab:gwi-yang}). While NeuSpeech \citep{yang2024neuspeech} shows promising character error rates, it does not surpass chance in BLEU or ROUGE. MAD \citep{yang2024mad} is similar, but with worse character errors. NeuGPT \citep{yang2024neugpt} is better on n-gram overlap metrics, but has a character error rate worse than random. Moreover, all of these methods show BERTScores only on par with their random selection baselines. Thus, none of these approaches convincingly surpass chance across all metrics. Our methods show superior performance across the board.

\begin{table}
    \centering
    \begin{tabular}{l|llllll}
        \toprule
        \textbf{Method} & WER $\downarrow$ & CER $\downarrow$ & BLEU $\uparrow$ & ROUGE $\uparrow$ & METEOR $\uparrow$ & BERT $\uparrow$ \\
        \midrule
        Rand. selection & $1.00_{\pm .0002}$ & $.87_{\pm .001}$ & $.07_{\pm .002}$ & $.08_{\pm .002}$ & $.06_{\pm .001}$ & $.77_{\pm .0004}$ \\
        \midrule
        \textit{Noise inputs} \\
        Greedy & $1.00_{\pm .0003}$ & $.87_{\pm .004}$ & $.07_{\pm .0004}$ & $.07_{\pm .001}$ & $.05_{\pm .001}$ & $.74_{\pm .0003}$ \\
        \hspace{1em}+rand. fill & $1.00_{\pm .0003}$ & $.91_{\pm .004}$ & $.07_{\pm .0004}$ & $.08_{\pm .001}$ & $.05_{\pm .001}$ & $.77_{\pm .001}$ \\
        Beam & $0.99_{\pm .001}$ & $.85_{\pm .003}$ & $.10_{\pm .004}$ & $.10_{\pm .004}$ & $.06_{\pm .003}$ & $.74_{\pm .001}$ \\
        \hspace{1em}+rand. fill & $0.99_{\pm .001}$ & $.88_{\pm .003}$ & $.10_{\pm .004}$ & $.10_{\pm .004}$ & $.06_{\pm .003}$ & $.77_{\pm .001}$ \\
        \hspace{1em}+fill & $0.98_{\pm .001}$ & $.75_{\pm .006}$ & $.15_{\pm .008}$ & $.16_{\pm .008}$ & $.09_{\pm .006}$ & $.78_{\pm .003}$ \\
        \hspace{1em}+IC fill & $0.99_{\pm .0004}$ & $.77_{\pm .003}$ & $.16_{\pm .005}$ & $.16_{\pm .004}$ & $.10_{\pm .003}$ & $.79_{\pm .001}$ \\
        IC transcribe & $0.99_{\pm .001}$ & $.84_{\pm .004}$ & $.13_{\pm .001}$ & $.13_{\pm .001}$ & $.08_{\pm .002}$ & $.79_{\pm .001}$ \\
        \midrule
        \textit{Real inputs} \\
        Greedy & $\mathbf{0.90}_{\pm .002}$ & $.77_{\pm .002}$ & $.19_{\pm .003}$ & $.20_{\pm .003}$ & $.12_{\pm .002}$ & $.74_{\pm .001}$ \\
        \hspace{1em}+rand. fill & $\mathbf{0.90}_{\pm .002}$ & $.78_{\pm .002}$ & $.19_{\pm .003}$ & $.20_{\pm .003}$ & $.13_{\pm .003}$ & $.79_{\pm .001}$ \\
        Beam & $\mathbf{0.90}_{\pm .001}$ & $.76_{\pm .001}$ & $.22_{\pm .002}$ & $.22_{\pm .002}$ & $.13_{\pm .002}$ & $.75_{\pm .0003}$ \\
        \hspace{1em}+rand. fill & $\mathbf{0.90}_{\pm .002}$ & $.77_{\pm .002}$ & $.22_{\pm .002}$ & $.22_{\pm .002}$ & $.13_{\pm .002}$ & $.79_{\pm .0002}$ \\
        \hspace{1em}+fill & $\underline{0.91}_{\pm .002}$ & $\mathbf{.69}_{\pm .001}$ & $\mathbf{.25}_{\pm .002}$ & $\mathbf{.25}_{\pm .002}$ & $\mathbf{.15}_{\pm .002}$ & $\underline{.80}_{\pm .001}$ \\
        \hspace{1em}+fill (OOV-D) & $\underline{0.91}_{\pm .002}$ & $\mathbf{.69}_{\pm .001}$ & $\underline{.24}_{\pm .002}$ & $\mathbf{.25}_{\pm .002}$ & $\mathbf{.15}_{\pm .001}$ & $\underline{.80}_{\pm .001}$ \\
        \hspace{1em}+IC fill & $\mathbf{0.90}_{\pm .002}$ & $\underline{.71}_{\pm .002}$ & $\underline{.24}_{\pm .003}$ & $\underline{.24}_{\pm .003}$ & $\mathbf{.15}_{\pm .003}$ & $\mathbf{.81}_{\pm .001}$ \\
        IC transcribe & $\mathbf{0.90}_{\pm .004}$ & $.75_{\pm .004}$ & $.21_{\pm .005}$ & $.22_{\pm .005}$ & $.14_{\pm .004}$ & $\underline{.80}_{\pm .001}$ \\
        \bottomrule
    \end{tabular}
    \caption{\textbf{Our method is significant against all critical random baselines.} Random selection picks a  word uniformly randomly from the story vocabulary. Noise inputs indicates when random noise of the same mean and standard deviation as true samples is input to the trained model. We use a retrieval set of 250 words and decode the LibriBrain dataset for all experiments in this table.}
    \label{tab:ablations}
\end{table}

An ablation over our approach (Table \ref{tab:ablations}) shows that it is highly significant, surpassing random selection and random noise baselines across all metrics unlike prior attempts at B2T. We are careful to provide random noise baselines as they verify that our model does not overfit to the target story and that the results are attributable to decoding brain data. \citet{jo2024eeg} showed that this is a critical failing of prior work. When using real MEG inputs, we note that our beam method significantly improves n-gram metrics. Even more significant is beam+fill, which improves almost everything. Beam+IC fill performs similarly and is excellent for semantics (BERTScore) which we conjecture is because it has the entire sequence as context. Our method performs well even without annotated OOV positions, underscoring the strength of our OOV detection approach. With 88\% AUROC, the detector reliably distinguishes neural responses to in-vocabulary and OOV words, enabling accurate automatic in-filling. IC transcribe is also effective, though does not outperform true rescoring.

\subsection{Overcoming the Data Bottleneck}
\label{sec:data}

\begin{figure}
    \begin{center}
        \includegraphics[width=\linewidth]{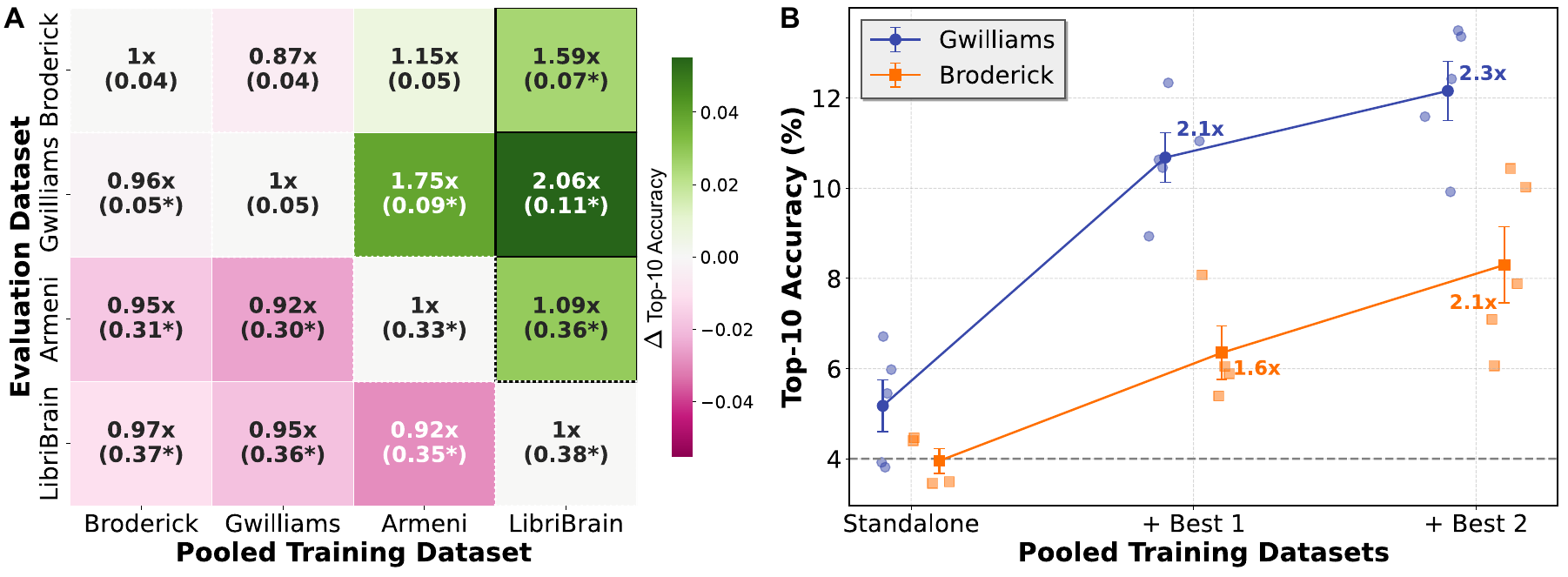}
    \end{center}
    \caption{\textbf{Selectively pooling data improves accuracy by $\mathbf{2.3\times}$.} \textbf{(A) Pairing datasets in training.} Numbers show accuracy improvement on the evaluation dataset when trained additionally with the training dataset. Numbers in brackets show raw accuracy (asterisks are statistically significant against chance). The diagonal shows no paired training, i.e. standalone. Shading shows the change in accuracy relative to standalone. \textbf{(B) Exploiting selective pooling doubles accuracy.} We combine the target data (Gwilliams and Broderick) with the best, and the best and second best, datasets by quality, i.e. LibriBrain and Armeni, leading to dramatic improvements on the target data.}
    \label{fig:data}
\end{figure}

On word classification accuracy, we show that selective pooling of datasets improves performance by more than double (Figure \ref{fig:data}). This also improves WER on B2T (\textit{pooled} in Table \ref{tab:gwi-yang}).

Speech decoding suffers from a data bottleneck as datasets are typically small and collected from few subjects. This bottleneck exists even for the largest MEG speech decoding datasets (Appendix \ref{app:scaling-bad}). 
Continued improvements from scaling indicate that more data could increase performance further.

One way to increase data is to combine datasets. This turns out not to be trivial. Since neural data come from heterogeneous sources with different numbers of sensors, we transform samples of various sensor dimensions to a consistent latent spatial dimension. We use a spatial attention \citep{defossez2023decoding} which projects samples using attention scores derived from the $(x, y)$ coordinates of each sensor. We note that there are many other plausible ways of doing this without using spatial information that perform equally well (Appendix \ref{app:positions}), mainly attributing improvements to selective pooling rather than any harmonisation method. We leave handling other sources of heterogeneity between datasets, such as anatomical, cognitive, and other hardware differences to the neural network.

While prior work has not seen significant improvements with pooled data in supervised decoding, we show for the first time that this is possible. Figure \ref{fig:data}A demonstrates that training with higher quality data, as measured by standalone performance, improves accuracy on lower quality datasets. In fact, the standalone performance of a dataset correlates strongly and significantly with the average improvement from jointly training with it ($r=.95$, $p=.048$). Even when jointly training EEG data with MEG data, as in the case of Broderick, we see that MEG, which is higher in quality, raises accuracy on the EEG data to statistically significant performance. This suggests that jointly training noisier modalities with higher quality data may improve accuracy on these modalities.

Combining multiple higher quality datasets with a target dataset leads to even more significant gains (Figure \ref{fig:data}B). Here, we again treat standalone performance as a measure of dataset quality and the metric by which to choose whether to pool a dataset with another. The factors that affect quality include the modality of the data, the number of subjects, the experimental protocol, the number of hours of recordings, and much more. We find that standalone performance is a useful overall proxy for quality as it takes into account all of these facets by being learned from the data.

\subsection{Scaling Vocabulary Size}

\begin{figure}
    \centering
    \includegraphics[width=1.0\linewidth]{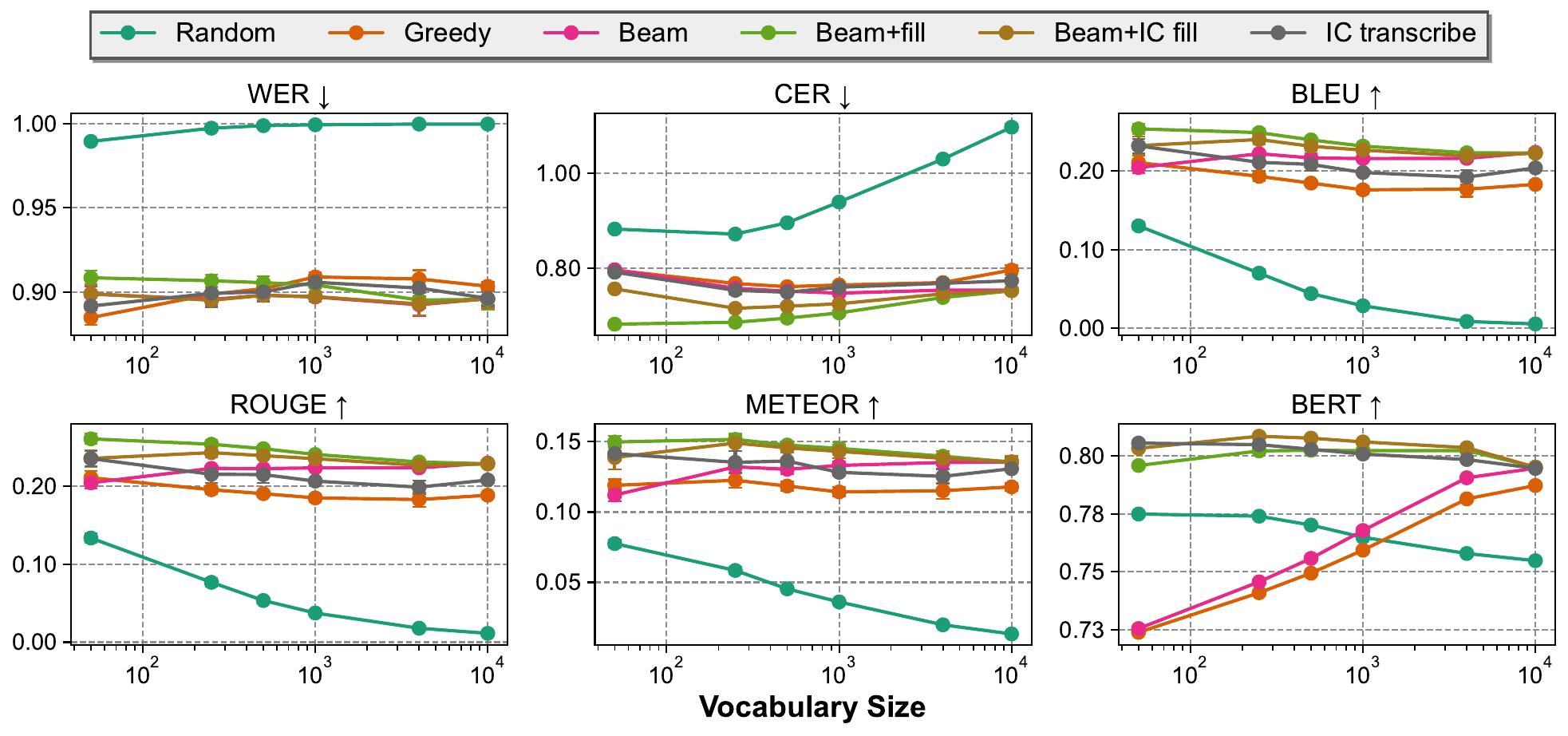}
    \caption{\textbf{Optimal vocabulary size differs by method and metric.} We train our model with increasingly large vocabularies and show the effect on each decoding strategy's performance. Methods with/without in-filling tend to converge with larger vocabularies as there are fewer words to in-fill.}
    \label{fig:vocab-scaling}
\end{figure}

Increasing the vocabulary size lowers the number of OOV predictions but reduces the test accuracy within-vocabulary, even on a fixed set of words (Appendix \ref{app:vocab-coverage}). When leveraging sentence decoding strategies, this trade-off implies there is an optimal vocabulary size in which it is small enough for high word accuracy, but large enough that we do not need to in-fill too many words. Figure \ref{fig:vocab-scaling} reveals that the optimal vocabulary size actually differs for different methods and evaluation metrics. 

For WER, a smaller vocabulary is best, ensuring high word accuracy. Any signal on additional words with a larger vocabulary is offset by the general reduction in word classification accuracy. This is because WER is a hard measure where any difference to the ground truth word counts as an incorrect prediction. CER is softer as it operates at the character level where similar words, in terms of characters, will count as partially correct. Consequently, it can improve with the signal provided by larger vocabularies. Here, for most methods, the optimal vocabulary is 250 words. %

On the remaining metrics, we see mixed effects. With BLEU, ROUGE, and METEOR, methods are optimal with either 50 or 250-word vocabularies, with minor differences. Generally, larger retrieval sets lead to worse performance. On BERT, in-filling methods are best with a 250-word vocabulary. Somewhat surprisingly, greedy and beam improve with vocabulary size. The additional signal from using a larger vocabulary, while reducing exact matches, may increase the overall number of semantically relevant words in the prediction, leading to this improvement in BERT score.

\section{Conclusion \& Future Work}

We have demonstrated the first non-invasive brain-to-text system that 
comprehensively outperforms all chance baselines critical for establishing non-invasive B2T. Moreover, it consistently surpasses existing methods across all standard metrics, improving BLEU scores by up to $2.6\times$. Our approach combines three key innovations: extending word-level decoding to sequence decoding with language model rescoring, transforming closed-vocabulary word-level classifiers into effectively open-vocabulary decoders with predictive in-filling, and unlocking scaling of B2T systems through a selective dataset pooling strategy that more than doubles word classification accuracy.

Building on these results and their limitations (Appendix \ref{app:limitations}), we identify three future research directions: (1) \text{cross-modal training} by pooling invasive and non-invasive recordings to improve non-invasive performance through transfer from invasive data; (2) \text{scaling} up datasets to find the boundaries of non-invasive B2T; and (3) \text{improving} practical B2T without requiring exact matches, e.g. by boosting the semantics and coherence of decoded sequences.

\clearpage
\begin{ack}
We thank Sumeet Motwani for insightful conversations about leveraging large language models, Brendan Shillingford for early conversations around dataset pooling, Miran Özdogan for reviewing a draft of this paper, and the rest of the PNPL team for their thoughtful feedback.

The authors would like to acknowledge the use of the University of Oxford Advanced Research Computing (ARC) facility in carrying out this work. \url{http://dx.doi.org/10.5281/zenodo.22558}. We are especially grateful to the ARC support team for their timely support as conference deadlines approached.

DJ is supported by an AWS Studentship from the EPSRC Centre for Doctoral Training in Autonomous Intelligent Machines and Systems (AIMS) (EP/S024050/1).
GL is supported by an EPSRC Studentship.
OPJ and the PNPL group are supported by the MRC (MR/X00757X/1), Royal Society (RG$\backslash$R1$\backslash$241267), NSF (2314493), NFRF (NFRFT-2022-00241), and SSHRC (895-2023-1022).
\end{ack}

\bibliography{neurips_2025}

\appendix

\clearpage

\section{Sensor Positions and Dataset Pooling}
\label{app:positions}

A key concern in pooling data is that of the differences in sensor geometry between scanners. For example, LibriBrain uses 306 sensors, the Armeni scanner uses 269 sensors, and the Gwilliams scanner has 208. In each dataset, the sensor geometry is different, with sensors placed in different configurations and therefore picking up signals from different parts of the brain. Explicitly leveraging sensor position information in the neural network to provide useful inductive biases is one way in which datasets could be pooled more effectively.

In Table \ref{tab:pooling-methods}, we compare the spatial attention (SA) method \citep{defossez2023decoding} to inserting zeroes to match spatial dimensions between datasets (\textit{padding}), dataset-conditional linear or convolutional projections to a shared dimension (\textit{gating}) \citep{jayalath2024brain}, as well as spatial attention followed by gating. In all of these methods, once the data is in the same space, we allow the neural network to resolve other data differences during training.

We note that padding and gating are not explicitly provided spatial information via sensor positions and in all cases, the performance differences are not significant. While spatial attention can improve individual dataset performance \citep[Table A.2]{defossez2023decoding}, our results indicate that it does not benefit aggregated performance. This means that either spatial differences between data can be harmonised through learning without explicit knowledge of sensor positions, or that the networks are not yet able to make proper use of this spatial information.

\begin{table}[ht]
\centering
\begin{tabular}{l|cc}
\toprule
\textbf{Method} & Top-10 Accuracy & Different to SA? (\textit{p}-value) \\ 
\midrule
SA + Gating & $.127_{\pm .007}$ & No ($.62$) \\ 
Gating & $.124_{\pm .005}$ & No ($.75$) \\ 
Padding & $.119_{\pm .005}$ & No ($.79$) \\ 
SA & $.122_{\pm .007}$ & — \\ 
\bottomrule
\end{tabular}
\caption{\textbf{Pooled dataset performance with different pooling methods.} We evaluate Gwilliams while jointly training it alongside Armeni and LibriBrain. We conduct a $t$-test against the spatial attention method and find that none of the alternatives perform differently to a statistically significant degree with five seeds. Uncertainty is standard error.}
\label{tab:pooling-methods}
\end{table}

\section{Relaxing Alignment and Decoding Without Alignment}
\label{sec:alignment}

Although aligning brain data to word onsets is possible by getting patients to space their attempted speech (e.g. \citet{moses2021neuroprosthesis}), decoding without alignment is useful as it permits natural speaking patterns. Here, we investigate reducing alignment and decoding without alignment altogether.

\begin{wrapfigure}[17]{r}{0.4\textwidth}
    \includegraphics[width=\linewidth]{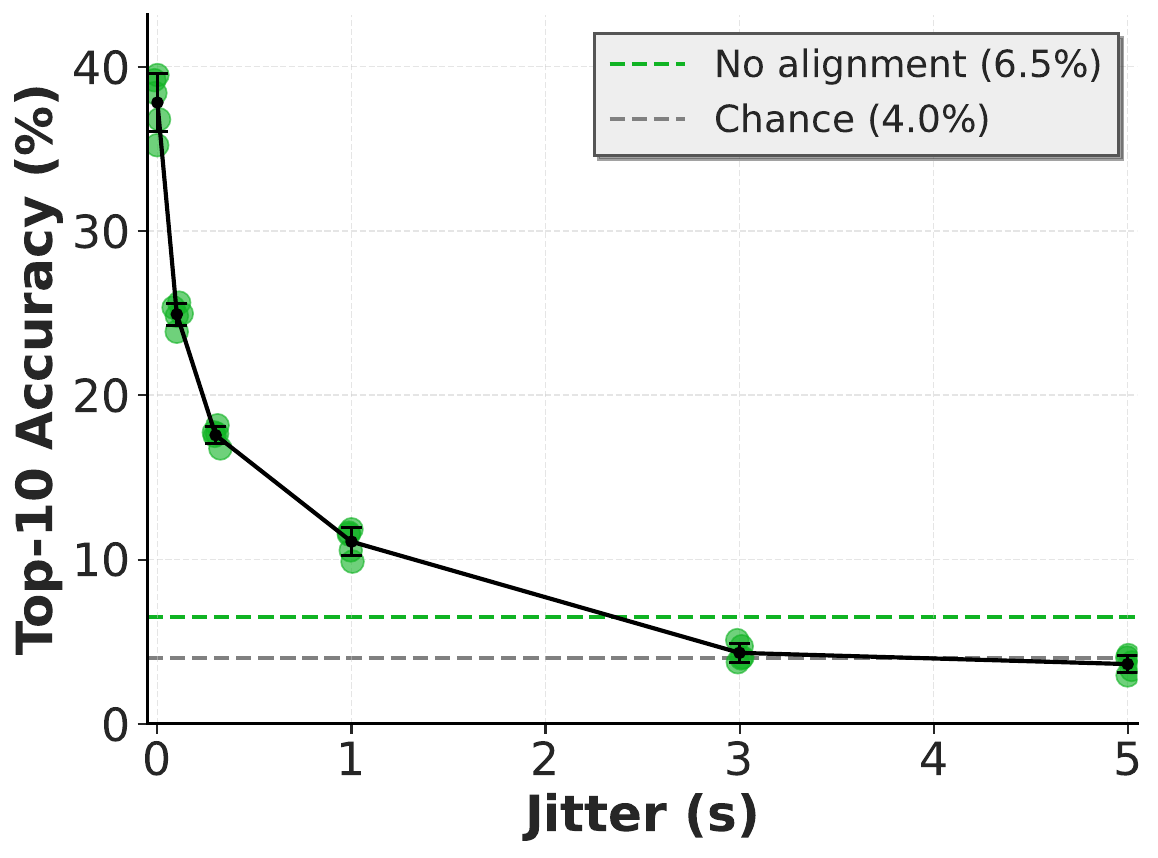}
    \caption{\textbf{Alignment is critical.} We add a random jitter in the range $[0, \mathrm{jitter}]$ to the aligned input samples. We train on LibriBrain with a vocabulary size of 250 and quote word classification accuracy.}
    \label{fig:alignment}
\end{wrapfigure}

Figure \ref{fig:alignment} shows that decoding is highly sensitive to alignment. Even a small jitter drastically reduces word classification accuracy. Still, removing alignment by evenly distributing the input samples in a sequence leads to statistically significant results.

However, evenly distributing inputs requires knowing the length of the target sequence. Thus, we use CTC-style merging, i.e. duplicate merging and alignment-free prediction, when even this is not known. This gives significantly better-than-chance CER, suggesting that this predicts meaningful acoustic signals (Table \ref{tab:ctc-full}).

Our merging method (a variant of \citet{graves2006connectionist}) decodes sequences from a trained model without known word onsets or alignments. In this setup, we space all the MEG windows at a fixed interval of 0.3s which is approximately the average time between words. We then make a set of predictions and slide all the segments until we reach the end of the sentence. The slide interval is half the MEG window spacing to make sure that we can reasonably resolve most words.

Once we have all the predictions, we average the probabilities for positions where there are multiple predictions and apply a 1D average pooling operation on the probabilities in the time dimension with a kernel size of 5 and stride of 3. Then, we collapse repeats in any section of consecutive repeats to a single prediction. CTC-greedy is then the argmax word predictions while CTC-beam emerges from applying LLM rescoring to reconstruct the sentence. All CTC results use a model trained on the LibriBrain dataset with a 250-word retrieval set.

\begin{table}
    \centering
    \begin{tabular}{l|llllll}
        \toprule
        \textbf{Method} & WER $\downarrow$ & CER $\downarrow$ & BLEU $\uparrow$ & ROUGE $\uparrow$ & METEOR $\uparrow$ & BERT $\uparrow$ \\
        \midrule
        Rand. selection & $\underline{1.00}_{\pm .0002}$ & $.87_{\pm .001}$ & $\underline{.07}_{\pm .002}$ & $\mathbf{.08}_{\pm .002}$ & $\mathbf{.06}_{\pm .001}$ & $\mathbf{.77}_{\pm .0004}$ \\
        CTC-Greedy & $\mathbf{0.99}_{\pm .001}$ & $\mathbf{.73}_{\pm .01}$ & $\underline{.07}_{\pm .003}$ & $.07_{\pm .002}$ & $.05_{\pm .002}$ & $.76_{\pm 001}$ \\
        CTC-Beam & $\underline{1.00}_{\pm .001}$ & $\underline{.74}_{\pm .011}$ & $\mathbf{.08}_{\pm .004}$ & $\mathbf{.08}_{\pm .005}$ & $\mathbf{.06}_{\pm .003}$ & $\mathbf{.77}_{\pm .004}$ \\
        \bottomrule
    \end{tabular}
    \caption{\textbf{CTC Decoding Results.}}
    \label{tab:ctc-full}
\end{table}

\section{Scaling Data Is Not Enough}
\label{app:scaling-bad}

\begin{figure}
    \centering
    \includegraphics[width=0.5\linewidth]{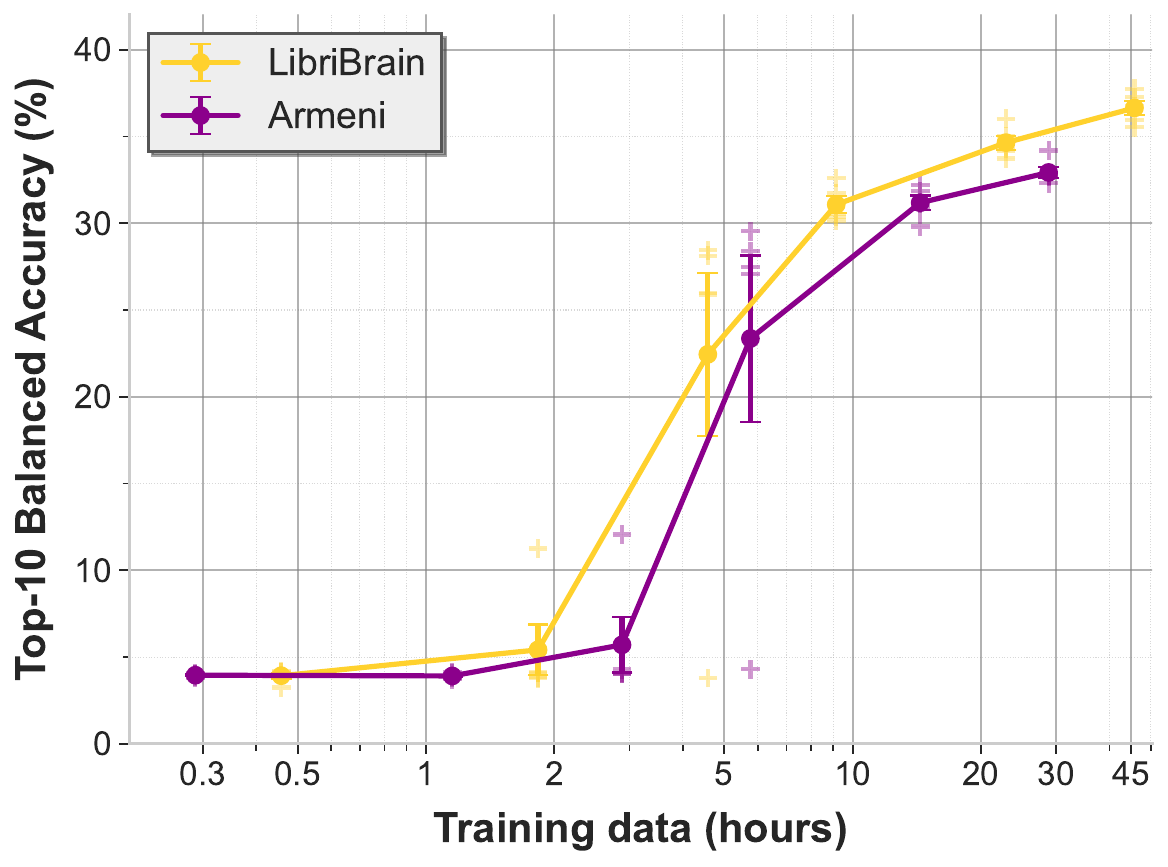}
    \caption{\textbf{Speech decoding is data-limited.} The plot shows top-10 word classification accuracy on the test set as we scale up the volume of data in the train set of LibriBrain and Armeni. To scale up, we randomly sample, without replacement, increasing numbers of samples from all sessions and subjects in the training set.}
    \label{fig:scaling}
\end{figure}

Among MEG speech decoding datasets, there are none with more than a hundred subjects, and none with more than 50 hours of within-subject recordings. As a result, speech decoding is data-limited, which we show in Figure \ref{fig:scaling}. Extrapolating the performance on LibriBrain suggests that scaling up to 80\% top-10 accuracy would require several orders of magnitude more data. Evidently, further work needs to be done on methods that scale in addition to collecting more data.

\section {Data Splits and Nonsense Correlations}
\label{app:data-nonsense}

We provide details on our data splits in Table \ref{tab:data-splits}. In designing our data splits, we were very careful to avoid nonsense correlations \citep{harris2020nonsense} by using independent sessions collected at different times. We also took care to avoid stimulus leakage within-dataset as well as across datasets when jointly training. This allows us to accurately evaluate content generalisation and ensures that all the information in the test set, including both the data and stimuli, are unseen. Taking random splits over these datasets, as is common in machine learning, would lead to leakage of stimuli as the same stimulus can be presented to multiple subjects. Even when stimuli are guaranteed to be non-overlapping, taking test set samples from the same recording as train set samples risks nonsense correlations as neural data and target text can be autocorrelated over time. For example, the semantics of two consecutive sentences in a session are likely to be very similar. Slow drift artifacts in the data at the time of these sentences can therefore be correlated with their semantics, thus leading to a nonsense correlation. 

\begin{table}[h]
    \centering
    \begin{tabular}{l|ccc}
        \toprule
        \textbf{Dataset} & Train & Validation & Test \\
        \midrule
        LibriBrain \citep{ozdogan2025libribrain} & Sherlock1-6 (excl. val and test) & Sherlock1 Sess. 9 & Sherlock1 Sess. 10-12 \\
        Armeni \citep{armeni202210} & Sess. 1-8 & Sess. 9 & Sess. 10 \\
        Gwilliams \citep{gwilliams2023introducing} & Task 2-3 & Task 0 & Task 1 \\
        Broderick \citep{broderick2018electrophysiological} & All sess. (excl. val and test) & Sess. 3, 9, \& 19 & Sess. 7, 8, \& 14 \\
        \bottomrule
    \end{tabular}
    \caption{\textbf{Data splits.} When training Armeni and LibriBrain jointly, we additionally exclude Sherlock3 session 9 and 10 from LibriBrain in the train set as the stories (The Adventure of the Engineer's Thumb and The Adventure of The Noble Bachelor) overlap with the validation and test set stories in the Armeni dataset. When comparing to the prior MEG brain-to-text work \citep{yang2024neuspeech, yang2024mad, yang2024neugpt}, we instead validate on Gwilliams task 1 and test on Gwilliams task 0 to match the experimental setup for the results quoted in \citet{yang2024neugpt}. We use all available subjects for train, validation, and test.}
    \label{tab:data-splits}
\end{table}

\begin{figure}
    \centering
    \includegraphics[width=0.5\linewidth]{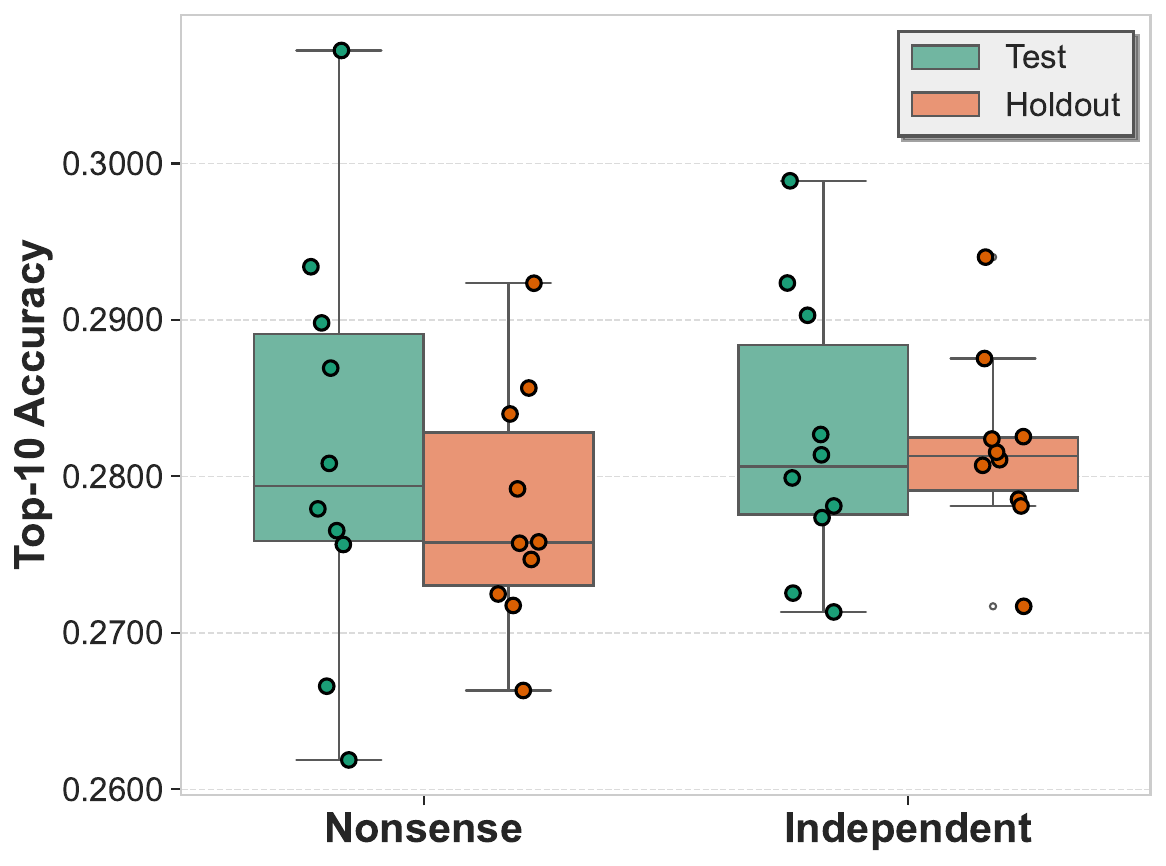}
    \caption{\textbf{Generalization with data splits containing overlapping vs independent sessions.} Refer to Table \ref{tab:nonsense} for complete details.}
    \label{fig:nonsense-box}
\end{figure}

In practice, we find that differences in results due to potential nonsense correlations are not statistically significant (Table \ref{tab:nonsense}). If there were nonsense correlations, we would expect the test set in the overlapping case to outperform the test set in the independent case due to overfitting to nonsense correlations. Similarly, we would expect the holdout set to perform worse than the test set in the overlapping case due to this overfitting. Here, and in Figure \ref{fig:nonsense-box}, we use the same experimental setup. We hypothesise that high-pass filtering, standardisation of samples, and baseline correction all help to avoid more significant nonsense correlations due to slow drift artifacts. Other sources of nonsense correlations may be insignificant in this experimental setup.

\begin{table}
    \centering
    \begin{tabular}{l|cc|c}
        \toprule
        \textbf{Evaluation set} & Overlapping & Independent & Significance \\
        \midrule
        Test     & $.282_{\pm .004}$ & $.282_{\pm .003}$ & n.s. ($p = .87$) \\
        Holdout  & $.278_{\pm .002}$ & $.282_{\pm .002}$ & n.s. ($p = .21$) \\
        \midrule
        Significance & n.s. ($p = .15$) & n.s. ($p = .77$) \\
        \bottomrule
    \end{tabular}
    \caption{\textbf{Generalisation with data splits of overlapping vs independent sessions.} We use subject 1 of the Armeni dataset for this experiment to avoid stimuli leakage. For independent session splitting, we randomly sample from the first nine sessions, seven sessions as the train set, and one session each as the validation and test set. For overlapping session splitting, we randomly split (without replacement) $\frac{7}{9}$ths of the data in the first nine sessions as train, another $\frac{1}{9}$th as validation, and the last $\frac{1}{9}$th as test. For both methods, we use session 10 as the holdout set to test generalisation beyond the potentially nonsensically correlated test set. We determine the significance of the difference between the data splitting approaches using 10 seeds. Note: n.s. = not significant.}
    \label{tab:nonsense}
\end{table}

\section{Vocabulary Scaling and Coverage}
\label{app:vocab-coverage}

\begin{figure}
    \centering
    \includegraphics[width=\linewidth]{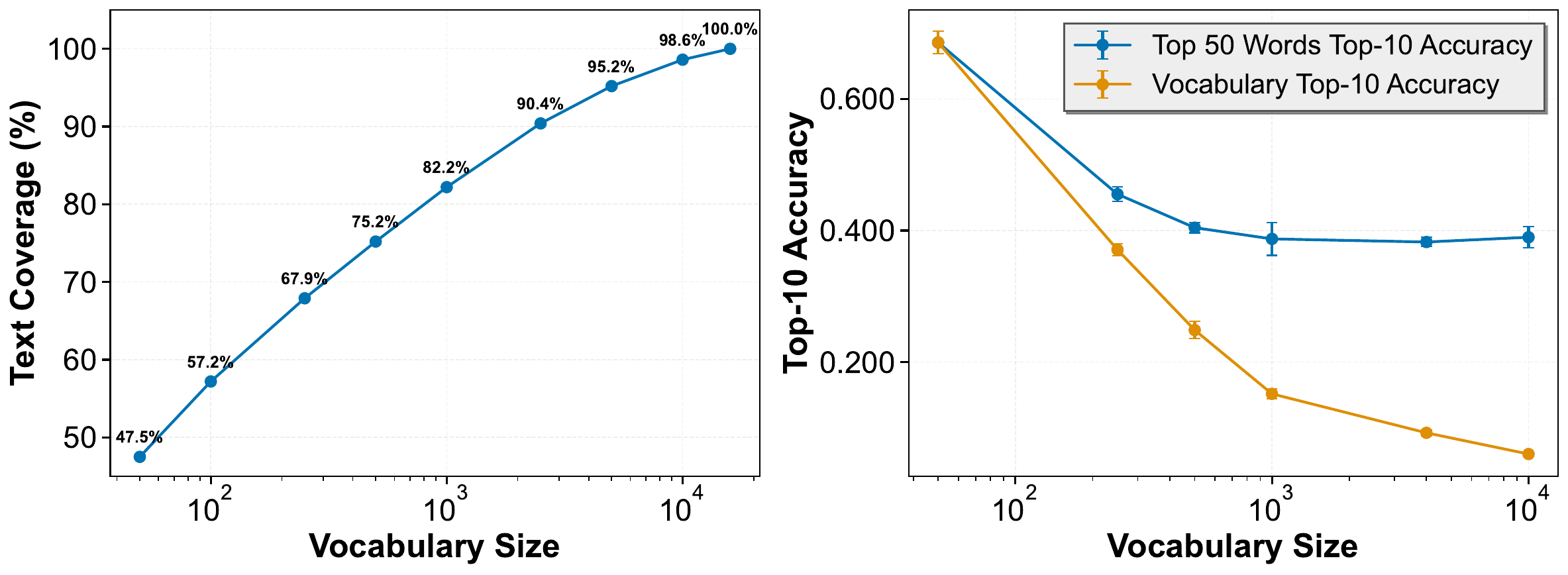}
    \caption{\textbf{Scaling vocabulary size on the LibriBrain dataset.} The left plot shows how much of the story a vocabulary of a certain size will cover. The right plot shows the top-10 word classification accuracy as we scale the vocabulary. The plot on the right supports a similar result seen by \citet[Figure 7]{dascoli2024decoding} in which scaling vocabulary size decreases accuracy.}
    \label{fig:vocab-coverage}
\end{figure}

In Figure \ref{fig:vocab-coverage}, on the left we show the percentage of the LibriBrain story text that can be covered by vocabularies of various sizes constructed from the $N$ most frequent words. Notably, a 50-word retrieval set covers 48\% of all words in the text and a 250-word dataset covers 68\%. Scaling up to 1000 words brings this up to only 82\%. On the right of the figure, we show that scaling up the retrieval set size reduces the decoding accuracy, first across the retrieval set, and second across a fixed set of the top 50 most frequent words, which are always in the vocabulary.

\section{Improving Evaluation With Missing Words}

\begin{figure}
    \centering
    \includegraphics[width=0.49\linewidth]{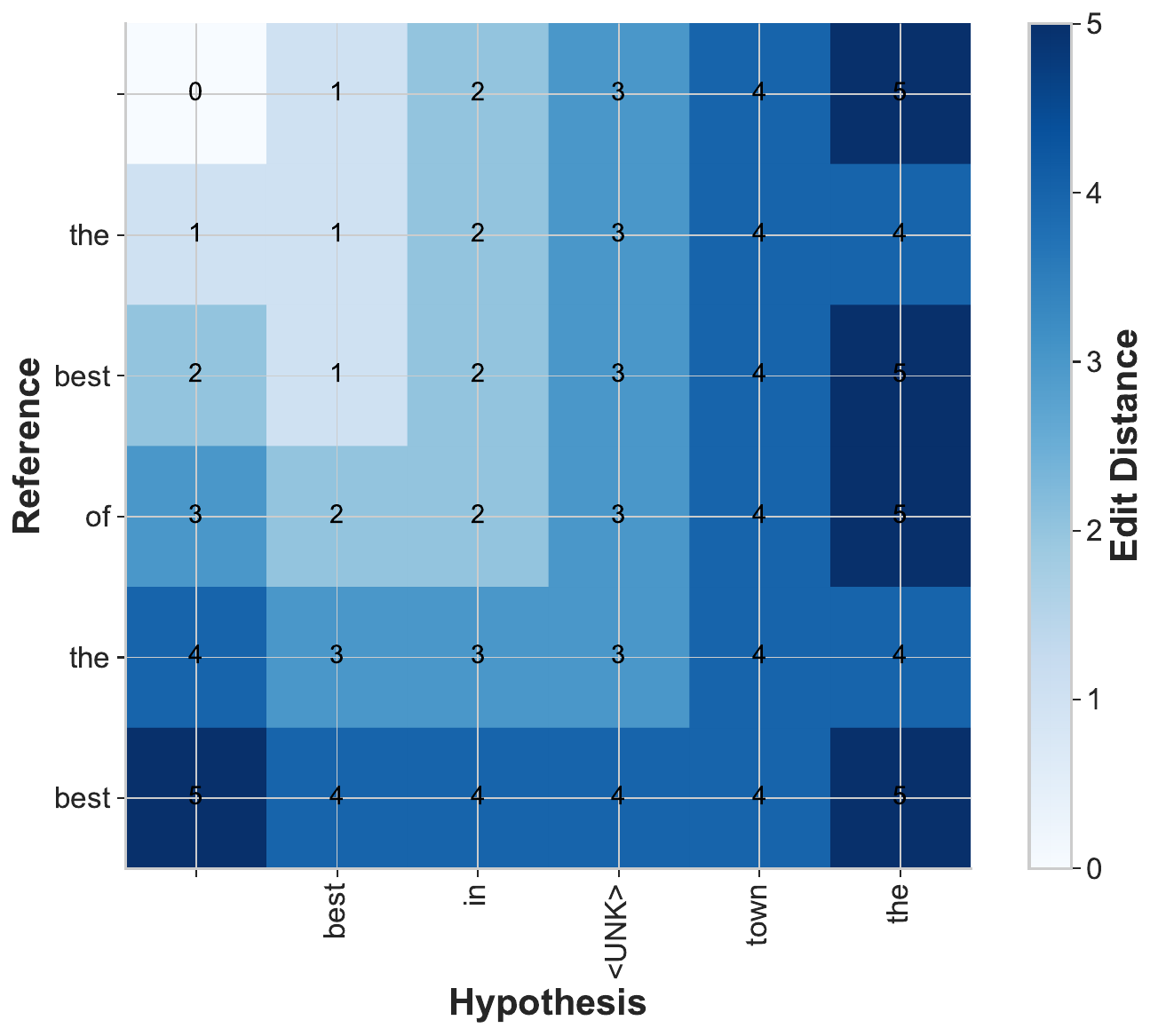}
    \includegraphics[width=0.49\linewidth]{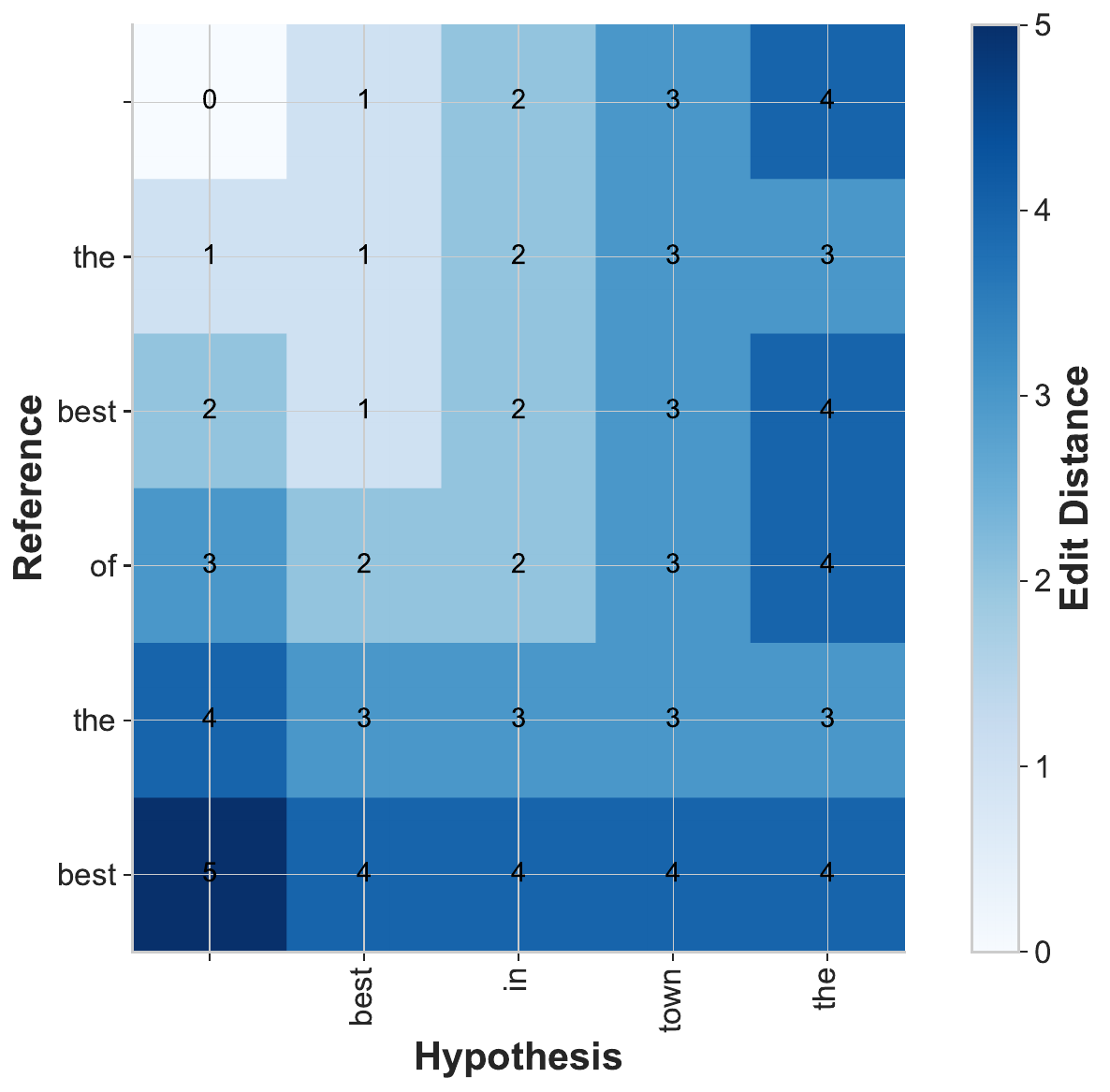}
    \caption{\textbf{Edit distance matrix with and without \texttt{<UNK>} for a toy example.}}
    \label{fig:better-eval}
\end{figure}

Excluding any text in place of out-of-vocabulary words can lead to artificial improvements in word error rates. This is caused by coincidental alignments when words are intentionally excluded. The following is a toy example where we include or exclude \texttt{<UNK>} at the position of an OOV word.

\begin{tcolorbox}
    \textbf{True sequence:}\\
    the best of the best\\\\
    \textcolor{black}{\textbf{Prediction}}:\\
    best in <UNK> town the\\
    \textit{Required edits: 5}\\\\
    \textcolor{black}{\textbf{Prediction (no <UNK>)}}:\\
    best in town the\\
    \textit{Required edits: 4}
\end{tcolorbox}

In the example, the same prediction but without an \texttt{<UNK>} token in place of the OOV position leads to a lower word error rate. This is because the lack of in-filling leads to a coincidental alignment of the word ``the'' (Figure \ref{fig:better-eval}). This implies an undesirable property in our evaluation: sentences with words filled in will not necessarily be \textit{at least as good} as sentences without any words filled in.

To address this, we always insert \texttt{<UNK>} tokens in OOV positions when no in-filling method is used. However, this can have its own side-effects, potentially affecting metrics other than WER. To ensure we have a fair evaluation, we also include in-filling with randomly selected words as a baseline in all of our method ablations.

\section{Additional Experiment Details and Hyperparameters}

\textbf{Statistical testing}\quad For all experiments, unless otherwise stated, we run 5 random seeds. Statistical tests are independent Welch's $t$-tests with the standard threshold for statistical significance ($p < .05$).

\textbf{Preprocessing}\quad We first notch filter the data around 50Hz to remove powerline artifacts. Then, we band-pass filter in the range 0.1-40Hz and resample the data to 50Hz. Following this preprocessing, we scale each recording such that the amplitude range $[-1, 1]$ covers the interquartile range and then clamp outliers above and below an amplitude of 5. Lastly, we apply a baseline correction to each individual sample, subtracting the mean of the sample calculated from the first 0.5s.

\begin{table}
    \centering
    \begin{tabular}{l|c}
        \toprule
        Hyperparameter & Value \\
        \midrule
        \textbf{Training} \\
        Batch size / seq. length & 64 \\
        Learning rate & 1e-5 \\
        Optimiser & AdamW \citep{adamw2019} \\
        Annealing schedule & Cosine (min. 1e-6 after 50 epochs) \\
        Early stopping patience & 5 epochs \\
        Early stopping metric & Val. top-10 word class. accuracy \\
        \midrule
        \textbf{Pooling} & Spatial attention \citep{defossez2023decoding} \\
        \quad Pooling output dim. & 270 \\
        \midrule
        \textbf{Signal encoder} & Brain model \citep{defossez2023decoding} \\
        Init. conv. & 1x1 (in: 270, out: 270) \\
        Subject layer & in: 270, out: 270 \\
        Main conv. channels & (270, 320, 320, ... x10)\\
        Kernel & 3 \\
        Stride & 1 \\
        Dilation growth & 2 \\
        Groups & 1 \\
        Dilation period & 5 \\
        \midrule
        \textbf{Context encoder} & \citet{dascoli2024decoding} \\
        Transformer depth & 16 \\
        Transformer heads & 16 \\
        Transformer dimension & 1024 \\
        Transformer attn. dropout & 0.1 \\
        Transformer pos. emb. & Rotary \\
        \midrule
        \textbf{Rescoring} \\
        Beam search rescorer & Llama 3.2 1B \\
        Rescorer weight ($\lambda$) & 1.5 \\
        Beam search in-filler & Llama 3.2 1B \\
        Max. words in context & 8 words \\
        Beam width & 5 beams \\
        \midrule
        \textbf{OOV position detector} & XGBoost \citep{chen2016xgboost} \\
        Features & Probs. + Table \ref{tab:oov-features} \\
        Learning rate & 0.05 \\
        No. of estimators & 200 \\
        Max. depth & 4 \\
        Min. child weight & 2 \\
        Subsample & 0.8 \\
        Subsample ratio columns by tree & 0.8 \\
        Gamma & 1 \\
        $\alpha$ (L1) & 0.1 \\
        $\lambda$ (L2) & 1 \\
        \midrule
        \textbf{LLM APIs} \\
        In-context in-filler & Claude 3.7 Sonnet \\
        & w/ 4096 thinking tokens \\
        In-context transcriber & Claude 3.7 Sonnet \\
        & w/ 4096 thinking tokens \\
        \bottomrule
    \end{tabular}
    \caption{\textbf{Hyperparameters.}}
    \label{tab:hparams}
\end{table}

\textbf{Extracting targets}\quad We optimise the transformer output embeddings for the target embeddings from the 12th layer of a pre-trained large T5 LLM. When words consist of multiple tokens, they are averaged into a single target embedding.

\textbf{In-filling during rescoring}\quad Although a nested beam search allows the in-filling method to guarantee complete word predictions, in practice, the highest probability predictions are generally singular tokens. Thus, to reduce inference time, we discard the nested beam search and use a single-token prediction instead.

\textbf{In-context transcription}\quad In our prompt (Appendix \ref{app:prompts} Box 2), we provide the probabilities of the top five predictions. Given a large retrieval set, these probabilities can be rather similar. To make these differences easier for an LLM to reason about, we sharpen the probabilities we provide by applying the softmax operator

\begin{equation}
    \label{eq:sharpen}
    \text{softmax}(z_i) = \frac{e^{z_i/0.01}}{\sum_{j=1}^{n} e^{z_j/0.01}},
\end{equation}

over the top five probabilities with a temperature of 0.01. We selected this temperature as it provided much better separation amongst the top probabilities.

\textbf{Compute resources}\quad Training a model on a single dataset takes up to 12 hours on an NVIDIA V100 GPU with 32GiB of GPU memory. We use 64GiB of system memory and require up to 200GiB of local storage. Generating sequences using beam search and other post-processing methods takes up to 4 hours. We ran all experiments on an internal cluster and estimate that overall, we used on the order of 1000 GPU hours. This includes preliminary experiments that did not make it into the paper.

\section{OOV Prediction}
\label{app:oov-pred}

After training our encoding model, we extract a feature vector for each encoder output from the training set containing: (1) the output probability distribution, computed via a softmax over the cosine similarities and (2) additional statistics calculated from the probability distribution (Table \ref{tab:oov-features}). We record whether each feature vector is associated with an OOV position or not and fit a binary classification XGBoost model \citep{chen2016xgboost} to the features. At test time, we use this model to select which words should be in-filled. This OOV detection method achieves 88\% AUROC on LibriBrain. Including the additional statistics as features leads to a small but meaningful improvement of 1-2\%.

An alternative method would be to train a binary OOV classification head on top of the transformer outputs. While also straightforward, this method entails re-training word classification models with OOV detection. Separating these concerns simplifies the process and allows OOV detectors to be reused with models trained on different seeds. Moreover, fitting an XGBoost model is much faster than training a word classifier from scratch.

\begin{table}
    \centering
    \begin{tabular}{l|p{9cm}}
        \toprule
        Feature & Description \\
        \midrule
        Entropy & Information-theoretic uncertainty measure: $-\sum(p_i \times \log_2(p_i))$ \\
        Variance & Statistical measure of dispersion in the data \\
        Mean & Average of all values \\
        Median & Middle value when sorted \\
        Max & Maximum value \\
        Min & Minimum value \\
        Skew & Measure of distribution asymmetry \\
        Kurtosis & Measure of peakedness/tail heaviness \\
        Gini & $1-\sum p_i^2$; measures impurity/diversity \\
        Top 1 prob. & Highest probability value \\
        Top 2 prob. & Second highest probability value \\
        Top 1 ratio & Ratio between highest and second highest: top1/top2 \\
        Peaks & Count of values above the mean \\
        Zeros & Count of near-zero values ($<10^{-10}$) \\
        Nonzeros & Count of non-zero values ($>10^{-10}$) \\
        90th percentile & 90th percentile value \\
        10th percentile & 10th percentile value \\
        90th / 10th percentile ratio & Ratio between 90th and 10th percentiles \\
        Top 5 sum & Sum of five highest probability values \\
        \bottomrule
    \end{tabular}
    \caption{\textbf{Additional descriptive features for OOV prediction.}}
    \label{tab:oov-features}
\end{table}

\section{Random Baselines}
\label{app:random-baselines}

For completeness, we define two kinds of random baselines. The first, which we call \textit{random selection}, is based on random chance given the vocabulary of the dataset. We generate our random selection baseline by uniformly randomly sampling from the decoder vocabulary for every in-vocabulary position and from the rest of the unique words in the dataset for every out-of-vocabulary position. This ensures a fair comparison for every in-vocabulary prediction and for every out-of-vocabulary in-filling step. The second random baseline we use is \textit{random noise} and inspired by the findings in \citet{jo2024eeg}. In this baseline, to test whether brain-to-text systems truly generalise or overfit to the stories in the dataset, we train a model as usual and then test it with random noise inputs of the same scale and variance as the MEG data. \citet{jo2024eeg} noticed that the models they tested this way failed to be any better than this baseline. Therefore, these models were not learning to use the brain data.

\section{Limitations}
\label{app:limitations}

As we establish the first above-chance electrophysiological non-invasive brain-to-text method, there remain some limitations which we hope future work will resolve to drive progress towards clinical application. Like the rest of the community, we use data from heard speech stimuli and not other forms of speech. The leap from perceived to inner or attempted speech is one that remains to be made among non-invasive approaches. Nevertheless, we expect the methods here to generalise to other varieties of speech.

Our dataset pooling framework improves performance through joint training with high-quality datasets. However, we have not shown the ability to improve overall performance across all datasets. Presently, this requires the collection of more and varied speech data. Still, jointly training EEG alongside MEG may make the decoding of speech from EEG, which does not require magnetic shielding, more feasible, opening the possibility of future mobile and low-cost speech BCIs.

Additionally, we rely on aligning neural data to word onsets. Although we make strides towards alignment-free decoding (Appendix \ref{sec:alignment}), further work is needed. However, decoding without alignment may not be necessary in a speech BCI as patients can space out the words that they attempt to speak (e.g. \citet{moses2021neuroprosthesis}) to naturally segment neural data by word onset.

The last and most important limitation is overall decoding performance. We outperform all existing methods and establish a significant baseline, but we do not see the present results as strong enough for a clinical speech BCI. Our view is that this is firstly a data problem, requiring scaling up data and collecting more datasets. Even though our scaling laws suggest that to achieve high word accuracy, much more data must be collected (Appendix \ref{app:scaling-bad}), we encourage finding methods that improve scaling efficiency and also propose re-evaluating what it means to achieve a realistic non-invasive speech BCI. The focus should be on practical necessity over exact decoding accuracy. While scaling up data $100\times$ may not bring non-invasive performance near invasive speech decoding word error rates, it could be enough to enable semantic decoding to a level where non-invasive speech BCIs are useful. As the barriers to surgical BCIs are so large, non-invasive BCIs can be clinically practical without achieving similar levels of performance.

\section{Broader Impacts}
\label{app:impact}

Although currently preliminary, maturity of research on non-invasive brain-to-text technology presents significant implications. Potential benefits include enabling communication for individuals with paralysis and advancing brain-computer interfaces generally. However, mature iterations of this technology raise concerns about brain data privacy, potential misuse of neural information, and accessibility disparities that may exacerbate existing inequalities. At present, we limit potential misuse by decoding heard speech rather than inner speech. As the field advances to inner speech, we encourage researchers to engage with ethical principles and set common ethical standards.

\section{Prompts}
\label{app:prompts}

\begin{tcolorbox}[title=\center{\textbf{Box 1: In-Context In-Filling}}, breakable, boxrule=0.5mm]
\textbf{\textcolor{black}{Prompt}}\\
I have a noisy speech recognition system which predicts 64 words at a time. I am going to give you its predictions in an ordered list. [UNK] indicates that the target word for that position is out-of-vocabulary. \\

I want you to fill in any [UNK] positions with words that you think fit well in the sequence. Do not replace anything that is not [UNK]. Your output should be formatted as a Python dictionary mapping all 64 positions (0-indexed) to words, preserving the system's predictions and replacing any [UNK] with your suggestions. Do not output anything else. \\

Output example: \\
\{0: "don't", 1: "the", 2: "scowl", ..., 63: "if"\} \\

Predictions from speech recognition system: \\

0: the\\
1: <UNK>\\
2: sat\\
3: on\\
...\\
63: built\\\\

\textbf{\textcolor{black}{Answer}}

\{0: "the", 1: "cat", 2: "sat", 3: "on", ..., 63: "build"\} \\
\end{tcolorbox}

\begin{tcolorbox}[title=\center{\textbf{Box 2: In-Context Transcription}}, breakable, boxrule=0.5mm]
\textbf{\textcolor{black}{Prompt}}\\
I have a noisy speech recognition system which predicts 64 words at a time. I am going to give you its predictions in an ordered list of pairs (word, probability). For each position, I give you the top-5 word predictions ordered from most likely to least likely along with their probabilities. [UNK] indicates that the target word for that position is out-of-vocabulary. \\

I want you to predict the most likely sequence from this information, picking the words from the predictions for each position that go best together. Where there is an [UNK] I want you to replace it with your own prediction for a word that fits well. In places with no [UNK], your job is to just pick the best fitting word from the predictions list (do not use any other word). Your output should be formatted as a Python dictionary mapping all 64 positions (0-indexed) to words. Do not output anything else. \\

Output example: \\
\{0: "don't", 1: "the", 2: "scowl", ..., 63: "if"\} \\

Predictions from speech recognition system: \\

0: (the, 0.45), (he, 0.23), (she, 0.15), (i, 0.11), (car, 0.09) \\
1: <UNK> \\
2: (house, 0.15), (inn, 0.13), (in, 0.09), (new, 0.05), (nought, 0.01) \\
... \\
63: (built, 0.35), (with, 0.20), (love, 0.15), (of, 0.17), (my, 0.05) \\

\textbf{\textcolor{black}{Answer}}

\{0: "the", 1: "cat", 2: "in", ..., 63: "love"\} \\
\end{tcolorbox}

\section{Decoding Examples}
\label{app:examples}

\begin{figure}
    \centering
    \includegraphics[width=\linewidth]{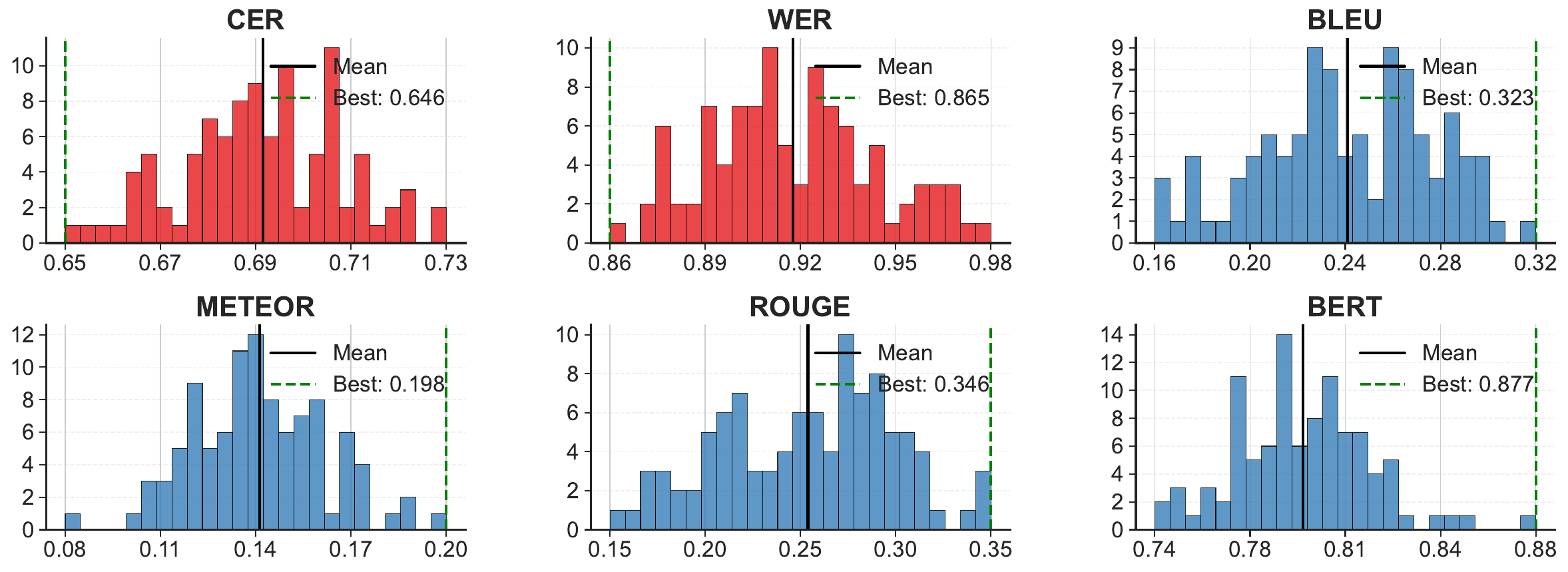}
    \caption{\textbf{Score histograms.} The scores are specific to the test set of LibriBrain for the experimental setup described in Appendix \ref{app:examples}.}
    \label{fig:hists}
\end{figure}

Examples in the table that follows are all generated using the rescoring and in-filling during beam search approach (beam+fill) from a single random seed on the LibriBrain dataset with a 250-word retrieval set. The table shows the best, median, and worst sequence for each metric. We also show the distributions of the scores for these predictions in Figure \ref{fig:hists}. 

Annotations have been automatically added to the examples table. \textcolor{teal}{\textbf{Exact matches}} are computed from a short exact match window where words are in similar locations between the stimulus and prediction. \textcolor{violet}{\underline{Similar n-grams}} are computed over words longer than two letters and where the \texttt{SequenceMatcher} from Python's \texttt{difflib} gives a word sequence similarity ratio greater than 0.9. Finally, \textcolor{blue}{\textit{semantically similar}} words are matched with SpaCy similarity with a threshold of 0.9. Words that match this way are annotated with a number to show their correspondence. We did not find any clear way to highlight semantic similarity between phrases when this is present.

\begin{longtable}{p{0.4\textwidth}p{0.4\textwidth}p{0.08\textwidth}}
\toprule
\textbf{Stimulus} & \textbf{Prediction} & \textbf{Score} \\
\midrule

\multicolumn{3}{c}{\textbf{Metric: WER}} \\
\midrule
\multicolumn{3}{c}{\textit{Best sequence}} \\
\midrule
at the thought of \textcolor{violet}{\underline{his}} \textcolor{violet}{\underline{own}} impotence what \textcolor{violet}{\underline{was}} \textcolor{violet}{\underline{that}} in the silence he \textcolor{blue}{\textit{heard}}[1] a gentle scratching sound low but very distinct in the quiet \textcolor{teal}{\textbf{\textcolor{teal}{\textbf{\textcolor{teal}{\textbf{\textcolor{teal}{\textbf{\textcolor{teal}{\textbf{\textcolor{teal}{\textbf{of the}}}}}}}}}}}} night it \textcolor{blue}{\textit{came}}[2] from \textcolor{teal}{\textbf{the \textcolor{teal}{\textbf{door \textcolor{teal}{\textbf{\textcolor{teal}{\textbf{of \textcolor{teal}{\textbf{\textcolor{teal}{\textbf{the house}}}}}}}}}}}} ferrier crept into the hall \textcolor{violet}{\underline{and}} \textcolor{blue}{\textit{listened}}[3] intently there \textcolor{violet}{\underline{was}} a pause \textcolor{violet}{\underline{for}} a few moments \textcolor{violet}{\underline{and}} then the low insidious sound \textcolor{violet}{\underline{was}} repeated someone \textcolor{violet}{\underline{was}} evidently & could be seen with our \textcolor{violet}{\underline{own}} eyes it is gone from the face \textcolor{violet}{\underline{that}} is the face \textcolor{teal}{\textbf{\textcolor{teal}{\textbf{\textcolor{teal}{\textbf{\textcolor{teal}{\textbf{\textcolor{teal}{\textbf{\textcolor{teal}{\textbf{of the}}}}}}}}}}}} earth face \textcolor{blue}{\textit{turned}}[1] towards to the north \textcolor{teal}{\textbf{\textcolor{teal}{\textbf{of \textcolor{teal}{\textbf{\textcolor{teal}{\textbf{the house}}}}}}}} \textcolor{violet}{\underline{and}} \textcolor{blue}{\textit{turned}}[1] out \textcolor{teal}{\textbf{the \textcolor{teal}{\textbf{door \textcolor{teal}{\textbf{\textcolor{teal}{\textbf{of \textcolor{teal}{\textbf{\textcolor{teal}{\textbf{the house}}}}}}}}}}}} to find out \textcolor{violet}{\underline{his}} father \textcolor{blue}{\textit{said}}[3] he \textcolor{violet}{\underline{was}} he \textcolor{violet}{\underline{was}} the son of a young man \textcolor{violet}{\underline{and}} \textcolor{violet}{\underline{that}} he had been in their employ \textcolor{violet}{\underline{for}} three years & 0.7969 \\
\midrule
\multicolumn{3}{c}{\textit{Median sequence}} \\
\midrule
\textcolor{violet}{\underline{the}} saints worn \textcolor{violet}{\underline{and}} exhausted he leaned upon \textcolor{violet}{\underline{his}} rifle \textcolor{violet}{\underline{and}} shook \textcolor{violet}{\underline{his}} gaunt \textcolor{violet}{\underline{hand}} fiercely at \textcolor{violet}{\underline{the}} silent widespread city beneath him as he looked at it he observed \textcolor{violet}{\underline{that}} there \textcolor{violet}{\underline{were}} flags in some of \textcolor{violet}{\underline{the}} \textcolor{blue}{\textit{principal}}[1] streets \textcolor{violet}{\underline{and}} other signs of festivity he was still speculating as to what this might mean \textcolor{violet}{\underline{when}} he heard \textcolor{violet}{\underline{the}} clatter of horse's hoofs \textcolor{violet}{\underline{and}} & of all \textcolor{violet}{\underline{the}} good people i know through my work with \textcolor{violet}{\underline{the}} \textcolor{violet}{\underline{his}} panic \textcolor{blue}{\textit{business}}[1] \textcolor{violet}{\underline{and}} with \textcolor{violet}{\underline{the}} \textcolor{violet}{\underline{his}} panic \textcolor{blue}{\textit{business}}[1] of us \textcolor{violet}{\underline{when}} we thought to us \textcolor{violet}{\underline{that}} we \textcolor{violet}{\underline{were}} we \textcolor{violet}{\underline{were}} going to get in \textcolor{violet}{\underline{the}} car \textcolor{violet}{\underline{and}} \textcolor{violet}{\underline{and}} so on for \textcolor{violet}{\underline{the}} it is not possible why many about these little things \textcolor{violet}{\underline{that}} we take a lot of for granted until & 0.9062 \\
\midrule
\multicolumn{3}{c}{\textit{Worst sequence}} \\
\midrule
led you safe to \textcolor{violet}{\underline{the}} chosen valley gave you a goodly share of \textcolor{violet}{\underline{land}} \textcolor{violet}{\underline{and}} allowed you to wax rich under our protection is not \textcolor{violet}{\underline{this}} so it is so \textcolor{blue}{\textit{answered}}[1] john ferrier in return for all \textcolor{violet}{\underline{this}} we asked but one condition that was that you should embrace \textcolor{violet}{\underline{the}} true faith \textcolor{violet}{\underline{and}} conform in every \textcolor{violet}{\underline{way}} to its usages \textcolor{violet}{\underline{this}} you promised to & question too broad in \textcolor{violet}{\underline{this}} case   as \textcolor{violet}{\underline{the}} \textcolor{blue}{\textit{question}}[1] is to find who is in \textcolor{violet}{\underline{the}} best position over their opponents while a few time on my head without any problem \textcolor{violet}{\underline{and}} i had my mind being \textcolor{violet}{\underline{away}} from long time now yes i have been trying to get \textcolor{violet}{\underline{this}} since  \textcolor{violet}{\underline{the}} whole end of \textcolor{violet}{\underline{this}} month \textcolor{violet}{\underline{and}} we are off & 1.0000 \\
\midrule
\multicolumn{3}{c}{\textbf{Metric: CER}} \\
\midrule
\multicolumn{3}{c}{\textit{Best sequence}} \\
\midrule
to keep to \textcolor{teal}{\textbf{the \textcolor{teal}{\textbf{right track}}}} for the moon had \textcolor{violet}{\underline{not}} yet risen \textcolor{teal}{\textbf{and the}} high cliffs on either \textcolor{violet}{\underline{side}} made the obscurity more profound weighed down with his burden and weary \textcolor{violet}{\underline{from}} his exertions he stumbled along keeping up his heart by the reflection that every step brought him nearer to lucy and that he carried with him enough to ensure \textcolor{violet}{\underline{them}} food & to be on \textcolor{teal}{\textbf{the \textcolor{teal}{\textbf{right track}}}} or the wrong was \textcolor{violet}{\underline{not}} so bad \textcolor{teal}{\textbf{and the}} right way of doing it about an hour or so ago out in my garden she was by my \textcolor{violet}{\underline{side}} but she was \textcolor{violet}{\underline{not}} in the mood to be nice to any one which will be of help and it will be of \textcolor{violet}{\underline{them}} before the lord \textcolor{violet}{\underline{from}} the & 0.5959 \\
\midrule
\multicolumn{3}{c}{\textit{Median sequence}} \\
\midrule
with bated breath lest something which \textcolor{blue}{\textit{fell}}[1] from \textcolor{violet}{\underline{their}} lips might be misconstrued \textcolor{violet}{\underline{and}} bring down a swift retribution upon them the victims of persecution had now turned persecutors on \textcolor{violet}{\underline{their}} own account \textcolor{violet}{\underline{and}} persecutors \textcolor{teal}{\textbf{\textcolor{teal}{\textbf{of the}}}} most terrible description not the inquisition of seville nor the german vehmgericht nor the secret societies of italy were ever able to put a \textcolor{violet}{\underline{more}} formidable machinery & \textcolor{blue}{\textit{cried}}[1] out in the morning having been out of bed about this hour \textcolor{violet}{\underline{and}} half into the night \textcolor{violet}{\underline{and}} after her it was a dream that it would be to make my life \textcolor{violet}{\underline{and}} my house a great place for both \textcolor{violet}{\underline{their}} children \textcolor{violet}{\underline{and}} \textcolor{violet}{\underline{their}} parents at the time \textcolor{teal}{\textbf{\textcolor{teal}{\textbf{of the}}}} birth \textcolor{teal}{\textbf{\textcolor{teal}{\textbf{of the}}}} child being far \textcolor{violet}{\underline{more}} than one of two or three & 0.6862 \\
\midrule
\multicolumn{3}{c}{\textit{Worst sequence}} \\
\midrule
led you safe to \textcolor{violet}{\underline{the}} chosen valley gave you a goodly share of \textcolor{violet}{\underline{land}} \textcolor{violet}{\underline{and}} allowed you to wax rich under our protection is not \textcolor{violet}{\underline{this}} so it is so \textcolor{blue}{\textit{answered}}[1] john ferrier in return for all \textcolor{violet}{\underline{this}} we asked but one condition that was that you should embrace \textcolor{violet}{\underline{the}} true faith \textcolor{violet}{\underline{and}} conform in every \textcolor{violet}{\underline{way}} to its usages \textcolor{violet}{\underline{this}} you promised to & question too broad in \textcolor{violet}{\underline{this}} case   as \textcolor{violet}{\underline{the}} \textcolor{blue}{\textit{question}}[1] is to find who is in \textcolor{violet}{\underline{the}} best position over their opponents while a few time on my head without any problem \textcolor{violet}{\underline{and}} i had my mind being \textcolor{violet}{\underline{away}} from long time now yes i have been trying to get \textcolor{violet}{\underline{this}} since  \textcolor{violet}{\underline{the}} whole end of \textcolor{violet}{\underline{this}} month \textcolor{violet}{\underline{and}} we are off & 0.7547 \\
\midrule
\multicolumn{3}{c}{\textbf{Metric: BLEU}} \\
\midrule
\multicolumn{3}{c}{\textit{Best sequence}} \\
\midrule
at the thought of \textcolor{violet}{\underline{his}} \textcolor{violet}{\underline{own}} impotence what \textcolor{violet}{\underline{was}} \textcolor{violet}{\underline{that}} in the silence he \textcolor{blue}{\textit{heard}}[1] a gentle scratching sound low but very distinct in the quiet \textcolor{teal}{\textbf{\textcolor{teal}{\textbf{\textcolor{teal}{\textbf{\textcolor{teal}{\textbf{\textcolor{teal}{\textbf{\textcolor{teal}{\textbf{of the}}}}}}}}}}}} night it \textcolor{blue}{\textit{came}}[2] from \textcolor{teal}{\textbf{the \textcolor{teal}{\textbf{door \textcolor{teal}{\textbf{\textcolor{teal}{\textbf{of \textcolor{teal}{\textbf{\textcolor{teal}{\textbf{the house}}}}}}}}}}}} ferrier crept into the hall \textcolor{violet}{\underline{and}} \textcolor{blue}{\textit{listened}}[3] intently there \textcolor{violet}{\underline{was}} a pause \textcolor{violet}{\underline{for}} a few moments \textcolor{violet}{\underline{and}} then the low insidious sound \textcolor{violet}{\underline{was}} repeated someone \textcolor{violet}{\underline{was}} evidently & could be seen with our \textcolor{violet}{\underline{own}} eyes it is gone from the face \textcolor{violet}{\underline{that}} is the face \textcolor{teal}{\textbf{\textcolor{teal}{\textbf{\textcolor{teal}{\textbf{\textcolor{teal}{\textbf{\textcolor{teal}{\textbf{\textcolor{teal}{\textbf{of the}}}}}}}}}}}} earth face \textcolor{blue}{\textit{turned}}[1] towards to the north \textcolor{teal}{\textbf{\textcolor{teal}{\textbf{of \textcolor{teal}{\textbf{\textcolor{teal}{\textbf{the house}}}}}}}} \textcolor{violet}{\underline{and}} \textcolor{blue}{\textit{turned}}[1] out \textcolor{teal}{\textbf{the \textcolor{teal}{\textbf{door \textcolor{teal}{\textbf{\textcolor{teal}{\textbf{of \textcolor{teal}{\textbf{\textcolor{teal}{\textbf{the house}}}}}}}}}}}} to find out \textcolor{violet}{\underline{his}} father \textcolor{blue}{\textit{said}}[3] he \textcolor{violet}{\underline{was}} he \textcolor{violet}{\underline{was}} the son of a young man \textcolor{violet}{\underline{and}} \textcolor{violet}{\underline{that}} he had been in their employ \textcolor{violet}{\underline{for}} three years & 0.4062 \\
\midrule
\multicolumn{3}{c}{\textit{Median sequence}} \\
\midrule
puzzled john ferrier sorely for his servants slept in an outhouse \textcolor{violet}{\underline{and}} the doors \textcolor{violet}{\underline{and}} windows \textcolor{violet}{\underline{had}} all been secured he crumpled the paper up \textcolor{violet}{\underline{and}} said \textcolor{teal}{\textbf{nothing to}} his daughter but the incident struck a chill into his heart the twenty nine \textcolor{violet}{\underline{days}} \textcolor{violet}{\underline{were}} evidently the balance \textcolor{teal}{\textbf{\textcolor{teal}{\textbf{\textcolor{teal}{\textbf{of the}}}}}} month which young \textcolor{violet}{\underline{had}} promised what strength or courage could avail against an enemy & question \textcolor{teal}{\textbf{\textcolor{teal}{\textbf{\textcolor{teal}{\textbf{of the}}}}}} day there was a time in my life \textcolor{violet}{\underline{where}} i really \textcolor{violet}{\underline{had}} \textcolor{teal}{\textbf{nothing to}} make any money these \textcolor{violet}{\underline{days}} the first thing is come clear from my mind \textcolor{violet}{\underline{and}} i can see the world before or after i go to sleep i have a dream that the world should open their eyes so that the whole \textcolor{teal}{\textbf{\textcolor{teal}{\textbf{\textcolor{teal}{\textbf{of the}}}}}} back \textcolor{teal}{\textbf{\textcolor{teal}{\textbf{\textcolor{teal}{\textbf{of the}}}}}} & 0.2344 \\
\midrule
\multicolumn{3}{c}{\textit{Worst sequence}} \\
\midrule
eked out by such employment as he could \textcolor{blue}{\textit{pick}}[1] up he travelled from \textcolor{violet}{\underline{town}} to \textcolor{violet}{\underline{town}} through \textcolor{violet}{\underline{the}} united \textcolor{violet}{\underline{states}} in quest of \textcolor{violet}{\underline{his}} enemies year passed into year \textcolor{violet}{\underline{his}} black hair turned grizzled but still he wandered on a human bloodhound with \textcolor{violet}{\underline{his}} mind wholly set upon \textcolor{violet}{\underline{the}} one object upon \textcolor{violet}{\underline{which}} he had devoted \textcolor{violet}{\underline{his}} life at last \textcolor{violet}{\underline{his}} perseverance was rewarded & \textcolor{violet}{\underline{the}} thing so far is that we do not doubt that \textcolor{violet}{\underline{the}} will of \textcolor{violet}{\underline{the}} people under \textcolor{violet}{\underline{the}} constitution of any \textcolor{violet}{\underline{state}} to be a \textcolor{violet}{\underline{state}} of their \textcolor{violet}{\underline{own}} there will be nothing for your course of action end of \textcolor{violet}{\underline{the}} \textcolor{blue}{\textit{day}}[1] in \textcolor{violet}{\underline{this}} case   about \textcolor{violet}{\underline{the}} same time \textcolor{violet}{\underline{which}} seemed to have been and again to him \textcolor{violet}{\underline{this}} \textcolor{blue}{\textit{day}}[1] by the & 0.1249 \\
\midrule
\multicolumn{3}{c}{\textbf{Metric: METEOR}} \\
\midrule
\multicolumn{3}{c}{\textit{Best sequence}} \\
\midrule
i have not married ferrier answered \textcolor{violet}{\underline{but}} women were few \textcolor{violet}{\underline{and}} there were \textcolor{violet}{\underline{many}} who had better claims than i i \textcolor{violet}{\underline{was}} not a lonely man i had my daughter to \textcolor{blue}{\textit{attend}}[1] to my wants it is of \textcolor{violet}{\underline{that}} daughter \textcolor{violet}{\underline{that}} i would speak to \textcolor{violet}{\underline{you}} said the leader \textcolor{teal}{\textbf{of the}} mormons she has \textcolor{blue}{\textit{grown}}[2] \textcolor{teal}{\textbf{to \textcolor{teal}{\textbf{be the}}}} flower of utah \textcolor{violet}{\underline{and}} has found & one of my best friends eyes my eyes my mind \textcolor{violet}{\underline{and}} i can see \textcolor{violet}{\underline{that}} in my mind of him i don't like her  thing \textcolor{violet}{\underline{but}} when he \textcolor{violet}{\underline{was}} the one in my life i thought \textcolor{violet}{\underline{was}} young \textcolor{violet}{\underline{and}} over the whole thing with me from the beginning \textcolor{teal}{\textbf{of the}} \textcolor{blue}{\textit{year}}[1] then this is \textcolor{teal}{\textbf{to \textcolor{teal}{\textbf{be the}}}} \textcolor{blue}{\textit{year}}[1] for \textcolor{violet}{\underline{you}} \textcolor{violet}{\underline{and}} \textcolor{violet}{\underline{many}} \textcolor{blue}{\textit{years}}[2] & 0.2389 \\
\midrule
\multicolumn{3}{c}{\textit{Median sequence}} \\
\midrule
suspense was unnerving he concealed \textcolor{violet}{\underline{his}} fears from \textcolor{violet}{\underline{his}} \textcolor{blue}{\textit{daughter}}[1] however \textcolor{violet}{\underline{and}} affected to make light of the whole \textcolor{blue}{\textit{matter}}[2] though \textcolor{violet}{\underline{she}} \textcolor{teal}{\textbf{\textcolor{teal}{\textbf{with the}}}} keen \textcolor{blue}{\textit{eye}}[3] of love \textcolor{violet}{\underline{saw}} plainly \textcolor{violet}{\underline{that}} he was \textcolor{violet}{\underline{ill}} at ease he expected \textcolor{violet}{\underline{that}} he would receive some message or remonstrance from \textcolor{violet}{\underline{young}} as to \textcolor{violet}{\underline{his}} conduct \textcolor{violet}{\underline{and}} he was not mistaken though it \textcolor{blue}{\textit{came}}[4] in an unlooked \textcolor{violet}{\underline{for}} & question of whether the patient \textcolor{violet}{\underline{will}} be \textcolor{teal}{\textbf{\textcolor{teal}{\textbf{with the}}}} same \textcolor{blue}{\textit{face}}[3] \textcolor{violet}{\underline{and}} body to go out \textcolor{teal}{\textbf{\textcolor{teal}{\textbf{with the}}}} \textcolor{violet}{\underline{young}} \textcolor{blue}{\textit{man}}[1] \textcolor{violet}{\underline{she}} \textcolor{violet}{\underline{saw}} over the counter at the pharmacy \textcolor{blue}{\textit{said}}[4] \textcolor{violet}{\underline{that}} we can get  our medication \textcolor{violet}{\underline{this}} \textcolor{blue}{\textit{way}}[2] is the most effective without causing or worsening \textcolor{violet}{\underline{and}} also \textcolor{violet}{\underline{that}} in other cases where it can be done while i find a few more \textcolor{violet}{\underline{for}} & 0.1408 \\
\midrule
\multicolumn{3}{c}{\textit{Worst sequence}} \\
\midrule
sacred council of four \textcolor{violet}{\underline{the}} \textcolor{blue}{\textit{girl}}[1] is young \textcolor{violet}{\underline{and}} we would \textcolor{violet}{\underline{not}} \textcolor{violet}{\underline{have}} her wed grey hairs neither would we deprive her of \textcolor{violet}{\underline{all}} choice we elders \textcolor{violet}{\underline{have}} \textcolor{violet}{\underline{many}} heifers1 but our children must also be provided stangerson has a son \textcolor{violet}{\underline{and}} drebber has a son \textcolor{violet}{\underline{and}} either of \textcolor{violet}{\underline{them}} would gladly welcome your daughter to their house let her choose between \textcolor{violet}{\underline{them}} they & question  to myself i \textcolor{violet}{\underline{have}} no doubt that i shall go back to it again \textcolor{violet}{\underline{and}} again through \textcolor{violet}{\underline{the}} years over \textcolor{violet}{\underline{the}} same period since  is quite large \textcolor{violet}{\underline{and}} can be \textcolor{violet}{\underline{all}} found in one place from \textcolor{violet}{\underline{the}} comfort \textcolor{violet}{\underline{and}} safety that will be so important to me \textcolor{violet}{\underline{any}} time i had to over his head it is \textcolor{violet}{\underline{not}} \textcolor{blue}{\textit{black}}[1] enough he & 0.0862 \\
\midrule
\multicolumn{3}{c}{\textbf{Metric: ROUGE}} \\
\midrule
\multicolumn{3}{c}{\textit{Best sequence}} \\
\midrule
at the thought of \textcolor{violet}{\underline{his}} \textcolor{violet}{\underline{own}} impotence what \textcolor{violet}{\underline{was}} \textcolor{violet}{\underline{that}} in the silence he \textcolor{blue}{\textit{heard}}[1] a gentle scratching sound low but very distinct in the quiet \textcolor{teal}{\textbf{\textcolor{teal}{\textbf{\textcolor{teal}{\textbf{\textcolor{teal}{\textbf{\textcolor{teal}{\textbf{\textcolor{teal}{\textbf{of the}}}}}}}}}}}} night it \textcolor{blue}{\textit{came}}[2] from \textcolor{teal}{\textbf{the \textcolor{teal}{\textbf{door \textcolor{teal}{\textbf{\textcolor{teal}{\textbf{of \textcolor{teal}{\textbf{\textcolor{teal}{\textbf{the house}}}}}}}}}}}} ferrier crept into the hall \textcolor{violet}{\underline{and}} \textcolor{blue}{\textit{listened}}[3] intently there \textcolor{violet}{\underline{was}} a pause \textcolor{violet}{\underline{for}} a few moments \textcolor{violet}{\underline{and}} then the low insidious sound \textcolor{violet}{\underline{was}} repeated someone \textcolor{violet}{\underline{was}} evidently & could be seen with our \textcolor{violet}{\underline{own}} eyes it is gone from the face \textcolor{violet}{\underline{that}} is the face \textcolor{teal}{\textbf{\textcolor{teal}{\textbf{\textcolor{teal}{\textbf{\textcolor{teal}{\textbf{\textcolor{teal}{\textbf{\textcolor{teal}{\textbf{of the}}}}}}}}}}}} earth face \textcolor{blue}{\textit{turned}}[1] towards to the north \textcolor{teal}{\textbf{\textcolor{teal}{\textbf{of \textcolor{teal}{\textbf{\textcolor{teal}{\textbf{the house}}}}}}}} \textcolor{violet}{\underline{and}} \textcolor{blue}{\textit{turned}}[1] out \textcolor{teal}{\textbf{the \textcolor{teal}{\textbf{door \textcolor{teal}{\textbf{\textcolor{teal}{\textbf{of \textcolor{teal}{\textbf{\textcolor{teal}{\textbf{the house}}}}}}}}}}}} to find out \textcolor{violet}{\underline{his}} father \textcolor{blue}{\textit{said}}[3] he \textcolor{violet}{\underline{was}} he \textcolor{violet}{\underline{was}} the son of a young man \textcolor{violet}{\underline{and}} \textcolor{violet}{\underline{that}} he had been in their employ \textcolor{violet}{\underline{for}} three years & 0.4062 \\
\midrule
\multicolumn{3}{c}{\textit{Median sequence}} \\
\midrule
\textcolor{violet}{\underline{when}} he \textcolor{blue}{\textit{heard}}[1] the click of the latch \textcolor{violet}{\underline{and}} looking through the window saw a stout sandy haired middle aged \textcolor{blue}{\textit{man}}[2] coming up the pathway \textcolor{violet}{\underline{his}} heart leapt to \textcolor{violet}{\underline{his}} mouth for this was none other than the great brigham young himself full of trepidation for he knew that such a visit boded him little \textcolor{blue}{\textit{good}}[3] ferrier ran to \textcolor{teal}{\textbf{the door}} to greet the & \textcolor{violet}{\underline{when}} i \textcolor{blue}{\textit{took}}[1] the test in the morning after my out \textcolor{teal}{\textbf{the door}} without a shower \textcolor{violet}{\underline{and}} a change of face cloth in the morning the first \textcolor{blue}{\textit{thing}}[3] to be done no course is a doubt \textcolor{violet}{\underline{and}} the same is between us \textcolor{violet}{\underline{and}} our children \textcolor{violet}{\underline{and}} we must have made a mess of \textcolor{violet}{\underline{this}} last point of mine on the \textcolor{blue}{\textit{death}}[2] of my own & 0.2500 \\
\midrule
\multicolumn{3}{c}{\textit{Worst sequence}} \\
\midrule
as keen as on \textcolor{violet}{\underline{that}} memorable \textcolor{blue}{\textit{night}}[1] when he \textcolor{violet}{\underline{had}} stood by john ferrier's grave disguised \textcolor{violet}{\underline{and}} \textcolor{violet}{\underline{under}} an assumed name he \textcolor{blue}{\textit{returned}}[2] to salt lake city careless what became of \textcolor{violet}{\underline{his}} \textcolor{violet}{\underline{own}} \textcolor{blue}{\textit{life}}[3] as long as he obtained what he knew to be justice \textcolor{violet}{\underline{there}} he found evil tidings awaiting him \textcolor{violet}{\underline{there}} \textcolor{violet}{\underline{had}} been a schism among \textcolor{violet}{\underline{the}} chosen people a few months & \textcolor{violet}{\underline{own}} ing \textcolor{violet}{\underline{that}} they are in business but they have done some of \textcolor{violet}{\underline{the}} best work \textcolor{violet}{\underline{here}} \textcolor{violet}{\underline{under}} \textcolor{violet}{\underline{the}} sun \textcolor{violet}{\underline{here}} we are in \textcolor{violet}{\underline{the}} middle of summer by now as we my mind \textcolor{violet}{\underline{and}} \textcolor{blue}{\textit{brought}}[2] up \textcolor{violet}{\underline{the}} subject with my \textcolor{violet}{\underline{head}} in my hands until i still don ’t have them i will find a \textcolor{blue}{\textit{way}}[1]  of making it in \textcolor{violet}{\underline{this}} \textcolor{blue}{\textit{world}}[3] & 0.1406 \\
\midrule
\multicolumn{3}{c}{\textbf{Metric: BERT}} \\
\midrule
\multicolumn{3}{c}{\textit{Best sequence}} \\
\midrule
that his wealth \textcolor{violet}{\underline{and}} position would be of no avail to him others as well known \textcolor{violet}{\underline{and}} as rich as himself had \textcolor{violet}{\underline{been}} spirited \textcolor{violet}{\underline{away}} before now \textcolor{violet}{\underline{and}} their goods given over \textcolor{teal}{\textbf{to the}} church he was a brave \textcolor{violet}{\underline{man}} \textcolor{violet}{\underline{but}} he trembled at the vague shadowy terrors which hung over him \textcolor{violet}{\underline{any}} known danger he could face with a \textcolor{blue}{\textit{firm}}[1] lip \textcolor{violet}{\underline{but}} this & you can use the same as what have \textcolor{violet}{\underline{been}} used for us to in \textcolor{violet}{\underline{any}} \textcolor{violet}{\underline{way}} to get in the door \textcolor{violet}{\underline{but}} those are about between them \textcolor{violet}{\underline{and}} they are not open \textcolor{teal}{\textbf{to the}} \textcolor{blue}{\textit{public}}[1] there is a small \textcolor{violet}{\underline{man}} who has \textcolor{violet}{\underline{been}} at my side since i came \textcolor{teal}{\textbf{to the}}se hands so long ago i am seen as a young \textcolor{violet}{\underline{man}} \textcolor{violet}{\underline{and}} i & 0.8256 \\
\midrule
\multicolumn{3}{c}{\textit{Median sequence}} \\
\midrule
vindictiveness far \textcolor{violet}{\underline{from}} doing so it \textcolor{violet}{\underline{had}} if anything augmented it the hunter's mind \textcolor{violet}{\underline{was}} of a hard unyielding nature \textcolor{teal}{\textbf{and the}} predominant \textcolor{blue}{\textit{idea}}[1] of revenge \textcolor{violet}{\underline{had}} taken \textcolor{violet}{\underline{such}} complete possession of it \textcolor{violet}{\underline{that}} \textcolor{violet}{\underline{there}} \textcolor{violet}{\underline{was}} no \textcolor{violet}{\underline{room}} for \textcolor{violet}{\underline{any}} other emotion he \textcolor{violet}{\underline{was}} however above \textcolor{violet}{\underline{all}} \textcolor{blue}{\textit{things}}[2] practical he soon \textcolor{blue}{\textit{realized}}[3] \textcolor{violet}{\underline{that}} even \textcolor{violet}{\underline{his}} iron constitution could \textcolor{violet}{\underline{not}} st\textcolor{teal}{\textbf{and the}} incessant strain \textcolor{violet}{\underline{which}} & question \textcolor{violet}{\underline{here}} \textcolor{violet}{\underline{from}} the \textcolor{blue}{\textit{point}}[1] we \textcolor{blue}{\textit{saw}}[3] in most of case to be case to be the case to be \textcolor{teal}{\textbf{and the}} be the \textcolor{blue}{\textit{like}}[4] of \textcolor{violet}{\underline{which}} is \textcolor{violet}{\underline{such}} as is in him \textcolor{violet}{\underline{that}} he can \textcolor{violet}{\underline{not}} see \textcolor{violet}{\underline{any}} light before him the whole \textcolor{violet}{\underline{room}} \textcolor{violet}{\underline{was}} \textcolor{violet}{\underline{his}} own and \textcolor{violet}{\underline{all}} he \textcolor{violet}{\underline{had}} to \textcolor{blue}{\textit{end}}[2] \textcolor{violet}{\underline{his}} own life and went to the police station where & 0.8024 \\
\midrule
\multicolumn{3}{c}{\textit{Worst sequence}} \\
\midrule
polygamy without a female population on which to draw was a barren doctrine indeed strange rumours began to be bandied \textcolor{violet}{\underline{about}} rumours of murdered immigrants \textcolor{violet}{\underline{and}} rifled camps in regions where indians had never been seen fresh women appeared \textcolor{teal}{\textbf{in the}} harems \textcolor{teal}{\textbf{\textcolor{teal}{\textbf{\textcolor{teal}{\textbf{of the}}}}}} elders women who pined \textcolor{violet}{\underline{and}} wept \textcolor{violet}{\underline{and}} bore upon their faces the traces of an unextinguishable horror belated wanderers upon & question \textcolor{violet}{\underline{about}} the use of i don't have time from the beginning \textcolor{teal}{\textbf{\textcolor{teal}{\textbf{\textcolor{teal}{\textbf{of the}}}}}} day to the \textcolor{teal}{\textbf{in the}} end end to an end to \textcolor{violet}{\underline{and}} end to with  \textcolor{violet}{\underline{and}} end to his own end \textcolor{teal}{\textbf{\textcolor{teal}{\textbf{\textcolor{teal}{\textbf{of the}}}}}} day the same day \textcolor{teal}{\textbf{\textcolor{teal}{\textbf{\textcolor{teal}{\textbf{of the}}}}}} week as the previous business day has closed out its business for today with an average of   is & 0.7696 \\
\midrule
\bottomrule
\end{longtable}

\end{document}